%% file: main.tex
\documentclass[10pt,a4paper,twocolumn]{article}
\usepackage{iftex}
\ifPDFTeX
  \usepackage[T1]{fontenc}
  \usepackage{lmodern}
\else
  \usepackage{fontspec}
\fi
\usepackage[margin=20mm,columnsep=6.3mm]{geometry}
\usepackage[english]{babel}
\usepackage{amsmath,amssymb,amsfonts,latexsym,bm}
\usepackage{graphicx,textcomp,url}
\usepackage[table]{xcolor}
\usepackage{framed,multirow,chngcntr,placeins,nicefrac}
\usepackage{algorithmic,tabu,bbm,diagbox,comment,array,makecell}
\usepackage{caption,tablefootnote,booktabs,float,siunitx,listings}
\usepackage{adjustbox,enumitem,longtable,tabularx,subcaption}
\usepackage[authoryear,round]{natbib}

\usepackage[hidelinks]{hyperref}
\usepackage{cleveref}
\renewcommand\arraystretch{1.22}
\title{AnyCXR: Human Anatomy Segmentation of Clinical Chest X-ray using Multi-stage Domain Randomized Synthetic Data with Imperfect Annotations and Conditional Joint Annotation Regularization Learning}
\author{\parbox{0.96\textwidth}{\centering
Zifei Dong\textsuperscript{1,2,*},
Wenjie Wu\textsuperscript{3,4,*},
Jinkui Hao\textsuperscript{1},
Tianqi Chen\textsuperscript{1},
Ziqiao Weng\textsuperscript{1,5},
Bo Zhou\textsuperscript{1,\textdagger}\\[6pt]
{\small
\textsuperscript{1}Department of Radiology, Northwestern University, Chicago, IL, USA\\
\textsuperscript{2}Data Science Institute, Vanderbilt University, Nashville, TN, USA\\
\textsuperscript{3}Department of Orthopedics, The Second Hospital of Shanxi Medical University, Taiyuan, Shanxi, P.R. China\\
\textsuperscript{4}Second Clinical Medical College, Shanxi Medical University, Taiyuan, Shanxi, P.R. China\\
\textsuperscript{5}School of Computer Science, University of Sydney, Sydney, NSW, Australia\\[5pt]
\textsuperscript{*}These authors contributed equally to this work.\\
\textsuperscript{\textdagger}Corresponding author: \href{mailto:bo.zhou@northwestern.edu}{bo.zhou@northwestern.edu}\\
\href{mailto:zifei.dong@vanderbilt.edu}{zifei.dong@vanderbilt.edu}
}}}
\date{}

\begin{document}
\maketitle

\begin{abstract}
\begingroup
Robust anatomical segmentation of chest X-rays (CXRs) remains challenging due to limited large-scale annotations, projection ambiguity, and substantial variability in real-world acquisition conditions. We propose \textbf{AnyCXR}, a unified framework for generalizable multi-anatomical segmentation across clinically relevant CXR acquisition views, including standard PA/AP, lateral, and moderate oblique projections, using synthetic supervision only. AnyCXR combines two complementary components. First, a \textbf{Multi-stage Domain Randomization (MSDR)} pipeline generates over 100,000 anatomically grounded and diverse synthetic radiographs from 3D CT volumes by introducing staged perturbations to projection geometry and image appearance. Second, a \textbf{Conditional Joint Annotation Regularization (CAR)} learning framework enables effective training from incomplete and imperfect CT-derived labels by enforcing conditional anatomical consistency in a shared latent space. Without using real-image supervision, models trained with AnyCXR generalize to multiple real-world CXR datasets and produce segmentation maps that can be efficiently refined into expert-quality annotations. The framework supports the delineation of 54 anatomical structures across PA, lateral, and oblique views. We further demonstrate that the resulting anatomical representations improve downstream applications, including automated cardiothoracic ratio estimation, scoliosis assessment, and CXR disease classification. These results suggest that synthetic radiography, combined with robust learning under imperfect supervision, provides an effective route toward scalable and anatomically informed CXR analysis across diverse imaging conditions.
\endgroup
\end{abstract}


\input{introduction}

\input{related_work}

\input{method}

\input{experiment}

\input{discussion}

\input{conclusion}

\section*{CRediT authorship contribution statement}
\textbf{Zifei Dong}: Writing – original draft, Visualization, Validation, Software, Project administration, Methodology, Investigation, Conceptualization. 
\textbf{Wenjie Wu}: Data curation, Formal analysis, Resources. 
\textbf{Jinkui Hao}: Writing – review \& editing. 
\textbf{Tianqi Chen}: Data Collection, Visualization. 
\textbf{Ziqiao Weng}: Writing – review \& editing. 
\textbf{Bo Zhou}: Writing – original draft, review \& editing, Supervision, Project administration, Methodology, Investigation, Conceptualization.

\section*{Declaration of competing interest}
The authors declare that they have no known competing financial interests or personal relationships that could have appeared to influence the work reported in this paper.

\section{Declaration of generative AI technologies use in the manuscript preparation process}
\begingroup During the preparation of this work the authors used ChatGPT in order to polish and suggest wordings after the content was written in order to ensure a natural, scientific flow. All contents, including methodology, analysis, results, and interpretations, were written prior to any use of AI. No text, data, or citations were generated by the model. After doing so, the authors reviewed and edited the content as needed and take full responsibility for the content of the published article. \endgroup

%%Harvard
\bibliographystyle{plainnat}
\bibliography{refs}

\clearpage

\input{Appendix}

\end{document}

%% file: introduction.tex
\section{Introduction}\label{sec:introduction}

\begingroup Chest X-ray (CXR) is the most widely used medical imaging modality worldwide, with billions of examinations acquired annually \citep{ai_CXR_R}. Owing to its low cost, rapid acquisition, and low radiation exposure, CXR serves as a first-line imaging tool across cardiothoracic, pulmonary, and musculoskeletal applications and plays a central role in screening and monitoring diseases such as pneumonia, tuberculosis, and lung cancer \citep{CheXmask}. As clinical AI increasingly shifts from image-level prediction toward anatomy-aware reasoning and quantitative analysis, robust anatomical segmentation has become an important foundation for downstream tasks including disease localization, structural measurement, and computer-assisted diagnosis.

In cross-sectional imaging modalities such as Computed Tomography (CT) and Magnetic Resonance Imaging (MRI), standardized acquisition geometry and well-defined three-dimensional anatomical boundaries have enabled highly reliable deep learning (DL)-based segmentation systems, including CT TotalSegmentator and MRI TotalSegmentator \citep{TS,TS-MRI,nnunet}. These models have become widely adopted components of automated imaging pipelines and downstream clinical research workflows \citep{TS_DStream}. In contrast, equivalent progress has not been achieved in CXR. Radiographs collapse complex 3D anatomy into a single projection, resulting in severe structure overlap, ambiguous boundaries, and appearance changes caused by acquisition geometry and patient positioning \citep{shadowlight}. Consequently, current CXR segmentation methods remain largely limited to a small number of structures and standard acquisition views, with reduced robustness under real-world variability.

A central obstacle is the lack of large-scale, anatomically consistent CXR annotations. Manual annotation of radiographs is labor intensive and inherently ambiguous because visible image boundaries do not always correspond to true anatomical surfaces. Synthetic supervision through Digitally Reconstructed Radiographs (DRRs) offers a promising alternative by projecting annotated CT volumes into radiograph-like images together with dense anatomical labels. However, this introduces a new challenge: the CT-derived segmentations used to generate supervision are typically produced automatically \citep{TS} and are therefore incomplete and imperfect. As synthetic datasets scale to cover broader acquisition diversity, annotation noise and partial labels become unavoidable. This raises the key question motivating this work: \textit{how can we effectively learn a generalizable anatomical segmentation model from large-scale synthetic radiographs with incomplete, imperfect, and heterogeneous supervision?}
\begin{figure*}[!tb]
    \centering
    \includegraphics[width=0.95\linewidth]{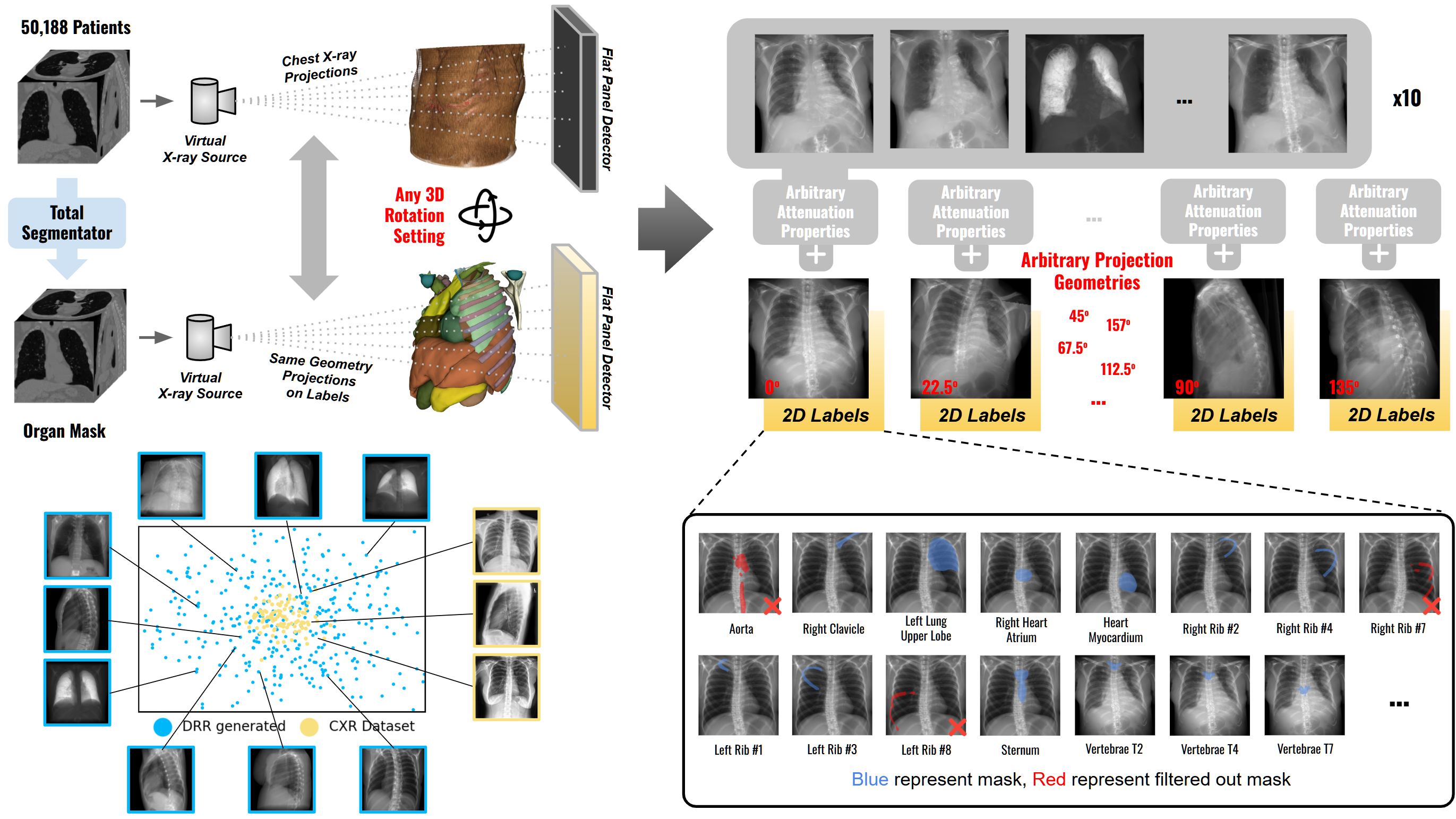}
    \caption{\small \textbf{AnyCXR Data Pipeline:} This workflow projects 3D CT volumes into diverse 2D Digitally Reconstructed Radiographs (DRRs) and segmentation masks using Multi-stage Domain Randomization. The pipeline includes an automated filtering step that discards invalid masks (red). The feature plot (bottom left) demonstrates that this method generates a data distribution significantly broader than typical real-world datasets.}
    \label{fig:Data}
    \vspace{-10pt}
\end{figure*}
To address this challenge, we propose \textbf{AnyCXR}, a framework for generalizable multi-anatomical segmentation of chest radiographs across clinically relevant acquisition views. AnyCXR integrates two complementary components. \textit{First}, we introduce a \textbf{Multi-stage Domain Randomization (MSDR)} pipeline for synthetic radiograph generation. Starting from 3D CT volumes, MSDR produces DRRs while introducing staged perturbations to projection geometry, tissue appearance, attenuation characteristics, and image formation conditions. This strategy substantially broadens the distribution of synthetic radiographs beyond conventional fixed-view rendering and enables the generation of more than 100,000 anatomically grounded CXRs spanning diverse projection angles and image appearances. To improve supervision quality, we further incorporate an automatic label-quality selection procedure that performs multi-level consistency checks and filters unreliable projected annotations. \textit{Second}, we propose \textbf{Conditional Joint Annotation Regularization (CAR)}, a learning framework designed for segmentation under incomplete supervision. CAR extends joint regularization to multi-anatomical segmentation by learning shared latent representations across structures and enforcing conditional consistency among anatomically related regions. This design allows the model to exploit both fully and partially labeled samples while preserving inter-structure coherence during training. Comprehensive experiments on synthetic and real-world datasets demonstrate that AnyCXR achieves robust anatomical segmentation across diverse acquisition conditions and clinically relevant projection views. We further show that the resulting anatomical representations improve multiple downstream applications, including cardiomegaly assessment, scoliosis measurement, and general CXR disease analysis. Our main contributions are summarized as follows:
\begin{itemize}
\vspace{-0.15cm}
\item We propose a Multi-stage Domain Randomization (MSDR) framework for generating large-scale synthetic chest radiographs with broad geometric and appearance diversity.

\vspace{-0.15cm}
\item We introduce an automatic label-quality selection strategy to enable robust learning from imperfect and partially labeled CT-derived supervision.

\vspace{-0.15cm}
\item We develop Conditional Joint Annotation Regularization (CAR), a segmentation framework that effectively leverages incomplete annotations while enforcing anatomical consistency.

\vspace{-0.15cm}
\item We demonstrate strong generalization across synthetic and real-world CXR datasets and show that the generated anatomical maps improve multiple downstream clinical AI tasks.
\end{itemize}

\endgroup

%% file: related_work.tex
\section{Related Work}\label{sec:related}

\subsection{Anatomical Segmentation of Chest Radiographs}

\begingroup Automatic anatomical segmentation of chest radiographs (CXRs) remains substantially more challenging than segmentation in cross-sectional imaging due to the projected nature of the modality and the limited availability of large-scale, anatomically consistent annotations. Existing CXR segmentation studies have primarily focused on a small number of structures, such as the lungs, heart, or clavicles, and are typically restricted to standard posteroanterior (PA) or anteroposterior (AP) acquisitions with manually curated labels \citep{CXRreview,PUNets}. Early approaches relied on traditional computer vision techniques, including edge detection, active contour models, and statistical shape priors \citep{features_seg,stats_seg,seg_review}. Although these methods can exploit local image gradients and anatomical regularities, they generally require extensive parameter tuning and are sensitive to acquisition variability, limiting robustness in heterogeneous clinical data. More recently, deep learning (DL) approaches trained on expert annotations have substantially improved segmentation accuracy and generalization \citep{CXR_review}. However, their performance remains fundamentally bounded by the quality and consistency of available annotations \citep{dual_decoder,partial_review}. Unlike volumetric modalities, anatomical boundaries in CXR are inherently ambiguous because projected structures overlap in 2D. Consequently, manual annotations often reflect visible contrast transitions rather than true anatomical surfaces \citep{Rib_seg}. For example, lung contours typically follow air–tissue interfaces and may not fully represent the underlying 3D anatomy extending behind the heart or diaphragm \citep{Lung_seg}. Similarly, structures such as the spine and ribs remain difficult to delineate consistently due to projection ambiguity. Generalization is further complicated by substantial acquisition variability. Even within standardized examinations, differences in patient positioning, respiratory state, beam angle, detector configuration, and field of view alter image appearance and effective anatomy projection \citep{XrayPosition}. As a result, models trained on narrow-view datasets often degrade when applied to oblique or non-standard projections \citep{cxrreview_4}. In contrast, CT provides standardized imaging geometry and anatomically consistent volumetric labels, enabling more reliable large-scale supervision \citep{CT_body}. This discrepancy motivates leveraging CT-derived information to improve robustness and anatomical fidelity in CXR segmentation \citep{CAD_AI}.\endgroup

\subsection{Synthetic Radiography and DRR-based Supervision}

\begingroup To address the scarcity and inconsistency of CXR annotations, recent work has explored generating synthetic radiographs from 3D CT volumes. By forward-projecting CT into digitally reconstructed radiographs (DRRs), synthetic datasets can be produced together with anatomically consistent pixel-level labels derived directly from volumetric segmentation. Several studies demonstrated the feasibility of this paradigm. SyntheX showed that models trained entirely on simple DRRs with image-level augmentation can achieve performance comparable to or exceeding models trained on real pelvic radiographs \citep{SynthX}. Similarly, CXAS applied CT-derived DRRs to chest imaging and reported strong agreement between predictions on synthetic and real CXRs \citep{CXAS}. These approaches are enabled by advances in DRR rendering. DeepDRR introduced a learning-based rendering framework that incorporates material decomposition, scatter simulation, and detector noise modeling to generate realistic radiographs from CT \citep{deepdrr}. DiffDRR later reformulated DRR generation as a differentiable projection operator, enabling efficient optimization and inverse problems such as 2D/3D registration \citep{diffdrr}. Together, these developments established DRR generation as a scalable and controllable source of supervision for radiographic learning. Synthetic radiography has subsequently expanded beyond data augmentation into large-scale representation learning. In interventional imaging, domain-randomized DRRs have enabled applications including surgical phase recognition \citep{pelphix} and autonomous X-ray acquisition systems \citep{killeen_autoxray}. More recently, FluoroSAM demonstrated that synthetic supervision can scale to foundation-model training through approximately three million synthetic X-rays with promptable segmentation capabilities across whole-body anatomy \citep{fluorosam}. In parallel, large-scale representation learning directly on real radiographs has also advanced chest imaging. Rad-DINO, for example, trained self-supervised image encoders on 838K chest radiographs and demonstrated strong transfer performance across downstream radiology tasks \citep{raddino,foundation}.

Despite these advances, existing synthetic radiography pipelines remain limited for comprehensive anatomical CXR segmentation. First, many prior studies focus on relatively constrained acquisition settings with limited geometric variation, reducing robustness to real-world projection diversity \citep{IF_DL_review}. Second, large-scale CT-derived labels generated automatically (e.g., using TotalSegmentator \citep{TS}) remain imperfect and introduce structure-dependent annotation noise into projected supervision. Our proposed AnyCXR framework extends prior DRR and domain-randomization paradigms toward dense multi-structure chest radiograph segmentation. Specifically, AnyCXR combines CT-derived supervision with CXR-oriented multi-stage geometric and appearance randomization, segmentation-aware quality control, and robust learning under imperfect projected labels to enable scalable segmentation across 54 anatomical classes and clinically relevant acquisition views.\endgroup

%% file: method.tex
% --- Custom Math Commands (Revised $\map$ Definition) ---
\newcommand{\vect}[1]{\mathbf{#1}}
\newcommand{\loss}[1]{\mathcal{L}_{\text{#1}}}
\newcommand{\R}{\mathbb{R}}
\newcommand{\volume}{\mathbf{V}} % The 3D volume (array)
\newcommand{\mask}[1]{\mathbf{M}_{\text{#1}}} % A general mask
\newcommand{\HU}{\text{HU}} % Hounsfield Unit
\newcommand{\dist}{\mathbf{D}} % Distance function or array
\newcommand{\I}{\mathbf{I}} % 2D Image
\newcommand{\map}[1]{\mathcal{T}_{\text{#1}}} % Transformation or mapping function

\begin{figure*}[htb!]
\begin{minipage}[b]{1.0\linewidth}
    \centering
    \centerline{\includegraphics[width=\columnwidth]{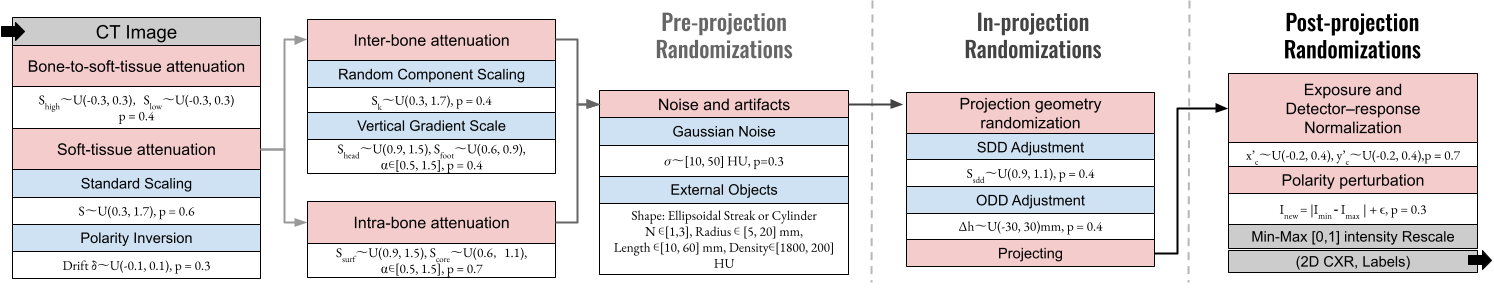}}
    
 \caption{\small{\textbf{MSDR Pipeline:} This flow chart details the Multi-stage Domain Randomization engine, which converts 3D CT data into 2D synthetic DRRs through a sequence of stochastic steps. The pipeline involves Pre-projection Randomizations acting on Hounsfield Units and image artifacts, In-projection Randomizations modifying geometry, and Post-projection Randomizations simulating detector response and inverting polarity. The red boxes denote randomization names, while the blue boxes denote different approaches.}}
	\label{fig:aug_parm}
 \end{minipage}
\end{figure*}

\section{Methods}
\label{sec:methods}
In this section, we provide an overview of the two core components of our framework and how they jointly enable robust CXR segmentation. First, we introduce the Multi-Stage Randomized Synthetic AnyCXR data-generation engine and the accompanying label quality-control pipeline, which together produce large, diverse datasets paired with high-quality partial organ annotations (Section~\ref{sec:engine}). Second, using these partially annotated datasets, we present the Conditional Joint Annotation Regularization (CAR) learning framework, which leverages partial labels and a latent-space regularization strategy to train a unified segmentation model capable of delineating all anatomical structures while preserving anatomical context (Section~\ref{sec:car}). Implementation details for both components are summarized in Section~\ref{sec:implementation}, and the study design and evaluation strategies are detailed in Section~\ref{sec:strategy}. 

\subsection{AnyCXR Generation Engine}
\label{sec:engine}

An overview of the AnyCXR generation engine is shown in Fig.~\ref{fig:Data}. The engine converts volumetric chest CT scans into radiograph-like projections, producing anatomically faithful, label-aligned synthetic CXRs for large-scale model training. The pipeline consists of two components: (1) segmentation-driven data curation to ensure anatomical plausibility and label reliability, and (2) a multi-stage randomized projection process that produces diverse CXRs paired with the curated labels.

\begin{center}
  \includegraphics[width=1.00\columnwidth]{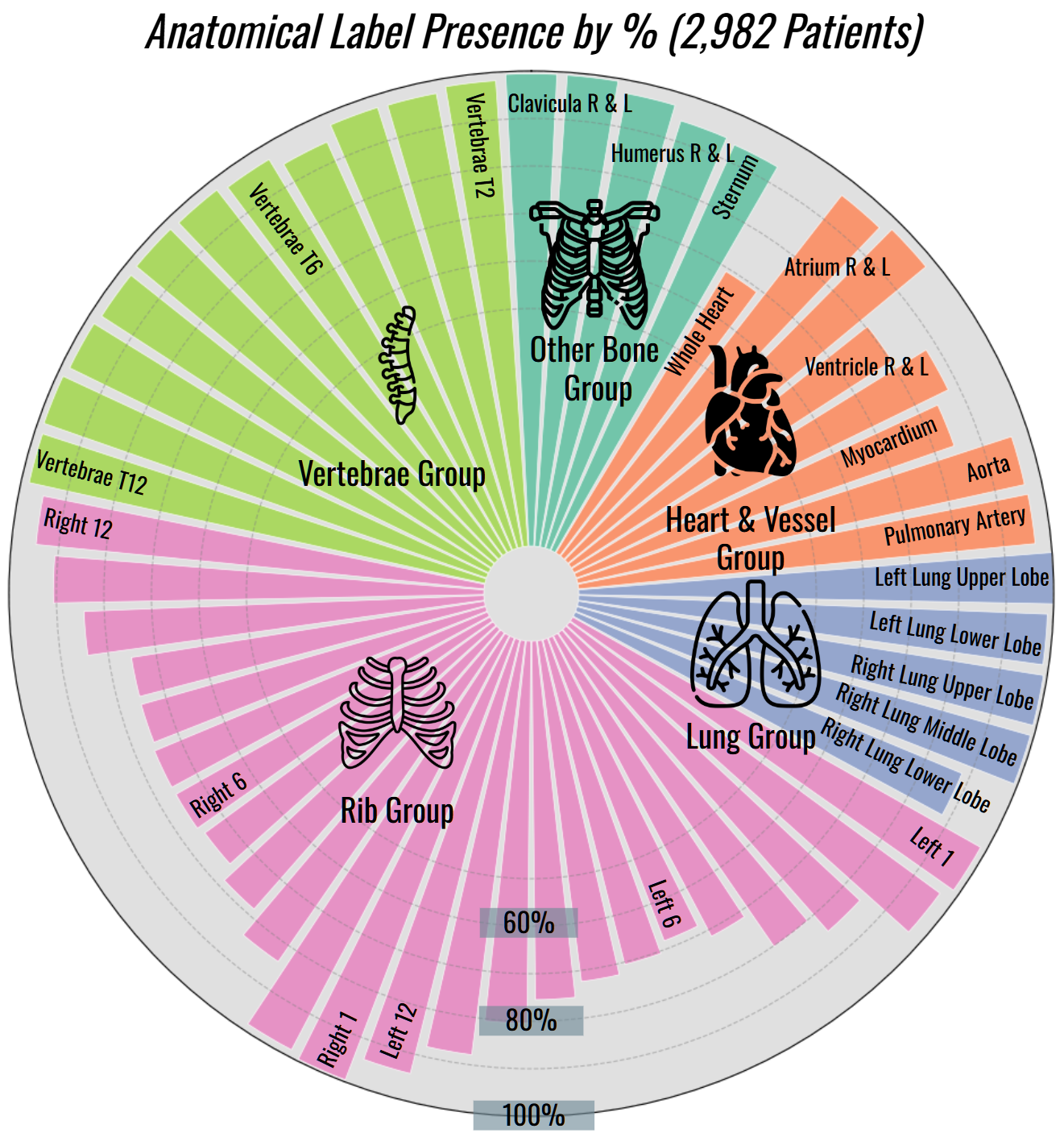}
  \captionof{figure}{\small{\textbf{Distribution of organ annotations after filtering:} This chart shows the anatomical label presence percentage across the 2,982 curated patient CT volumes after the two-stage quality control process. The distribution is grouped by major anatomical categories (Bone, Heart and Vessels, Lung, Rib, and Vertebrae), demonstrating the partial yet high-quality nature of the training dataset used for AnyCXR.}}
  \label{fig:dist1}
\end{center}

\subsubsection{Data Curation and Quality Control}

We begin by generating fine-grained anatomical masks for the CT-RATE dataset \citep{CTRate} (50{,}188 non-contrast chest CT scans from 21{,}304 patients) using CT TotalSegmentator (TS) \citep{TS}. Because the raw TS outputs exhibit variable reliability across this large and heterogeneous cohort, we introduce a two-stage segmentation-driven Quality Control (QC) procedure to ensure stable and trustworthy labels prior to projection.

\paragraph{Stage 1: 3D Anatomical Consistency Check}
This stage assesses the structural plausibility and integrity of each 3D mask to filter out volumes with severe segmentation failures. We focus on major skeletal structures, particularly the thoracic vertebrae, which serve as a sentinel for overall mask fidelity. For each CT volume, adjacent vertebral elements along the superior–inferior axis are examined slice-wise. A volume is discarded if (i) neighboring components exhibit substantial overlap or (ii) an implausible number of distinct elements appear within a single axial slice (e.g., three vertebrae in one slice, which is anatomically impossible).  
This 3D consistency rule removes the majority of gross segmentation errors. Volumes passing this check (2{,}982 CT scans) are retained for further QC and assumed to have predominantly reliable anatomy labels.

\paragraph{Stage 2: Per-Class Mask Reliability Check in 2D Projections}
The retained 3D masks are projected into DRRs at evenly spaced source–detector rotations from $0^{\circ}$ to $180^{\circ}$. For each view, we evaluate per-class mask reliability via a two-step process. First, isolated small components are removed by discarding connected components below a minimal fraction of the class’s largest component. If a second large component remains after cleanup, the class is flagged as unreliable and excluded from supervision for all projections of that CT.  
After this refinement, we retain a taxonomy of 54 anatomical classes, including cardiac chambers, great vessels, lung lobes, ribs (1–12), thoracic vertebrae (T2–T12), sternum, clavicles, and proximal humeri. Together, these QC procedures yield a curated dataset of 2{,}982 CT scans with high-quality partial segmentation labels suitable for downstream AnyCXR generation. The distribution of each organ is visualized in Fig.~\ref{fig:dist1} and the detailed numbers are in \textbf{\ref{app:C}}.

\subsubsection{Multi-stage Domain Randomization (MSDR)}
Standard DRR-based synthesis typically relies on 2D post-hoc augmentations (e.g., noise, brightness shifts). However, such operations do not capture the inherently 3D physical factors that drive appearance variation in real CXRs and beyond, such as bone-to-soft-tissue contrast under different exposures, heterogeneous bone mineralization, and complex tissue overlap. 

\begin{figure*}[htb!]
    \centering
    \centerline{\includegraphics[width=1.95\columnwidth]{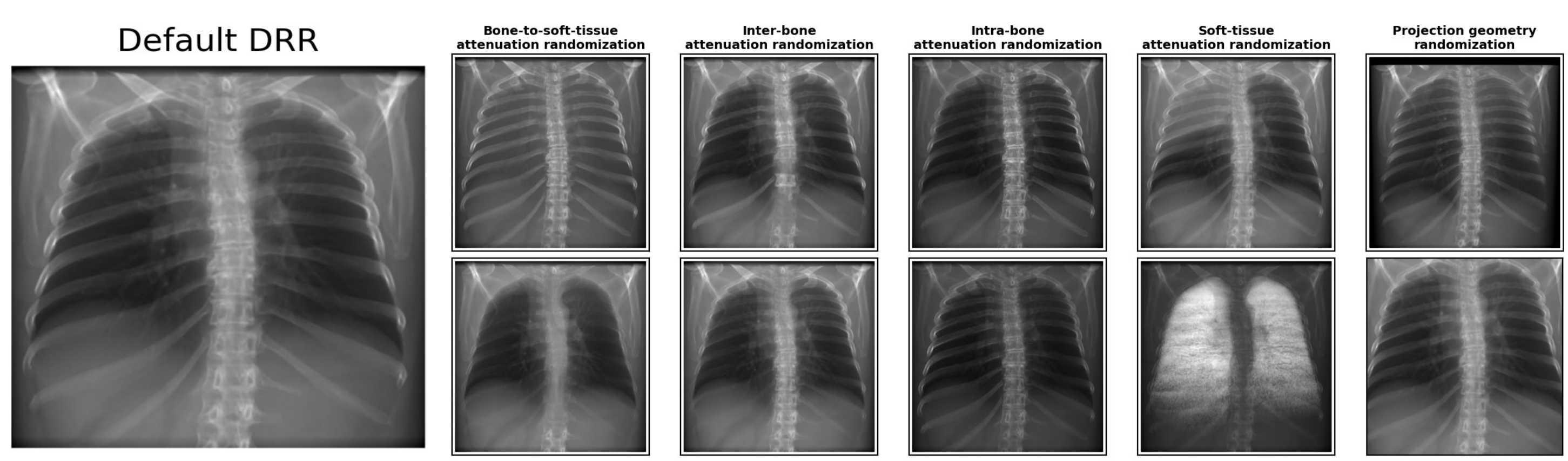}}
    
 \caption{\small{\textbf{DRR Augmentation Demonstration for Major Innovation:} This figure visually illustrates the effect of the primary Multi-stage randomization on the resulting DRR, starting from a Default DRR (left panel). Each subsequent column shows a different major 3D randomization applied in isolation: Bone-to-soft-tissue, Inter-bone, Intra-bone, Soft-tissue, and Projection geometry randomization, demonstrating how these 3D perturbations introduce diverse image appearances and contrast variations.}}
	\label{fig:aug_show}
\end{figure*}

To overcome these limitations, we propose Multi-stage Domain Randomization (MSDR). The workflow is illustrated in Fig.~\ref{fig:aug_parm} with technical details summarized in \textbf{\ref{app:A}}, inspired by Representation Learning \citep{anatomix} and AnyStar \citep{anystar}. Our MSDR engine performs the first stage of randomization directly in 3D \emph{before} projection. This “pre-projection" strategy perturbs the \emph{causes} of radiographic variability, producing a richer and more realistic distribution of DRRs. We introduce four principal 3D augmentation modules to modulate Hounsfield Units (HU), simulate structural variability, and alter projection behavior:
\begin{itemize}

    \vspace{-0.15cm}
    \item \textbf{Bone-to-soft-tissue attenuation randomization:} To replicate variations in bone–soft-tissue contrast arising from different exposure settings, we apply anti-correlated scaling factors to bone and soft tissue. Intensities are sampled from complementary intervals to disrupt consistent contrast relationships (e.g., attenuating bone signals while amplifying soft tissue).
    
    \vspace{-0.15cm}

    \item \textbf{Inter-bone attenuation randomization:} To enhance bone density heterogeneity, we independently perturb the attenuation of individual osseous components (e.g., vertebrae, ribs) to decouple adjacent structures, or introduce longitudinal intensity gradients to simulate varying anatomical thickness along the superior-inferior axis.
    
    \vspace{-0.15cm}

    \item \textbf{Intra-bone attenuation randomization:} To introduce realistic non-uniform bone density, we apply a morphology-based gradient that modulates attenuation as a function of voxel depth. This simulates the natural transition from the dense cortical surface to the trabecular core, creating heterogeneous internal textures.

    \vspace{-0.15cm}

    \item \textbf{Soft-tissue attenuation randomization:} To prevent the model from over-relying on absolute brightness for segmentation, we apply random multiplicative HU scaling perturbations to soft-tissue regions. Additionally, we introduce polarity inversion to simulate extreme contrast reversals, thereby generating a wider range of appearance variations.

    \vspace{-0.15cm}
    \item \textbf{Projection geometry randomization:} Source-to-Detector Distance (SDD) and Object-to-Detector Distance (ODD) are determined using available metadata when present. Otherwise, we use dataset-average values from CT-RATE and apply small random drifts to simulate variability in clinical setup.

    \vspace{-0.15cm}
    \item \textbf{Noise and artifacts:} 
    Gaussian noise is injected, and implant synthetic high-density objects (such as wires) are implanted to approximate common imperfections observed in clinical radiographs.

\end{itemize}

Following 3D augmentation, DRR projection is performed at a resolution of $512 \times 512$ using the modified CT intensities, and masks are propagated to all sampled view angles. After projection, in the second stage of 2D Detector-Level Randomization, we simulate detector physics and display characteristics using several 2D transformations:

\begin{itemize}
    \vspace{-0.15cm}
    \item \textbf{Exposure and Detector–response Normalization:} A randomized piecewise-linear window/level mapping is applied to reduce variations arising from exposure settings and display tone curves.
    \vspace{-0.15cm}
    \item \textbf{Polarity perturbation:} With low probability, image polarity is inverted (followed by re-normalization) to discourage reliance on absolute intensity polarity.
\end{itemize}

\begingroup
We additionally apply image domain 2D augmentations during training as part of our Detector-Level Randomization. These include horizontal and vertical flips, random rotations, additive Gaussian noise, simulated scatter artifacts, global contrast stretching, brightness perturbation, contrast perturbation, and gamma adjustment. 
\endgroup

Together, these pre-projection and post-projection randomizations form the MSDR engine, producing a highly diverse, physically plausible, and label-aligned synthetic CXR dataset for large-scale segmentation training. For a visual overview of the workflow logic and detailed implementation methods, please refer to Fig.~\ref{fig:aug_parm} and \textbf{\ref{app:A}}. We visualize the innovative 3D augmentation effect in Fig.~\ref{fig:aug_show}.

\begin{figure*}[t]
\begin{minipage}[b]{1.0\linewidth}
    \centering
    \centerline{\includegraphics[width=\columnwidth]{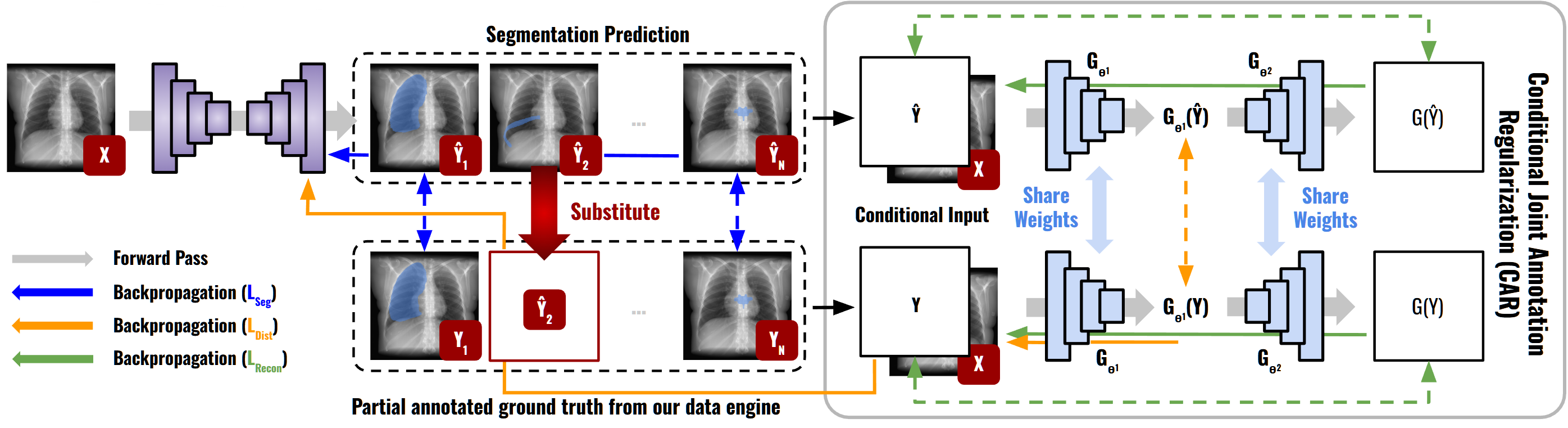}}
    
 \caption{\small{\textbf{CAR Framework Overview:} Input X-rays are segmented by a U-Net, with predictions substituting missing ground truth labels. Conditioned on image $\vect{X}$, these maps are encoded by $G_{\theta_1}$ to align their latent distributions via $\mathcal{L}_{\text{Dist}}$ (optimizing the U-Net), while the decoder $G_{\theta_2}$ ensures latent integrity via the reconstruction loss $\mathcal{L}_{\text{Recon}}$.}}
	\label{fig:Model}
 \end{minipage}
\end{figure*}

\subsection{Conditional Joint Annotation Regularization Model}
\label{sec:car}
The overall architecture is illustrated in Fig.~\ref{fig:Model}. Specifically, our segmentation model is built on a U-Net \citep{UNet} backbone with a ResNet50 encoder \citep{ResNet} pre-trained on ImageNet \citep{ImageNet}. Given an input image $\vect{X}$, the network $f_{\phi}$ outputs probabilistic segmentation maps $\hat{\vect{Y}} = f_{\phi}(\vect{X})$.

To handle partially labeled data, the segmentation loss $\mathcal{L}_{\text{Seg}}$ is computed selectively using a binary availability mask $\mathbf{M}$. Let $b$ denote the sample index within a batch of size $B$, and $c$ denote the class index. The entry $\mathbf{M}^{(b, c)}$ is set to 1 if annotations exist for class $c$ in sample $b$, and 0 otherwise. The loss is averaged over the available annotations:

\begin{equation}
    \mathcal{L}_{\text{Seg}} = \frac{1}{\sum \mathbf{M}} \sum_{b, c} \mathbf{M}^{(b, c)} \cdot \mathcal{L}_{\text{Dice}}(\hat{\vect{Y}}^{(b, c)}, \vect{Y}_{gt}^{(b, c)})
\end{equation}

While simple, this strategy fails to leverage anatomical relationships between organs.

To leverage cross-organ context even under partial labels, we introduce a conditional autoencoding module, termed \textit{Conditional Joint Annotation Regularization} (CAR). CAR enforces consistency between the predicted and anatomically plausible organ configurations, conditioned on the input image $\vect{X}$. CAR operates as an auxiliary autoencoder $G = G_{\theta_2} \circ G_{\theta_1}$ attached to the main segmentation head. Specifically, $G_{\theta_1}$ is the encoder that compresses the probability maps into a latent representation $\vect{z}$, which is used by the Distribution Loss ($\loss{Dist}$). $G_{\theta_2}$ is the decoder that reconstructs the maps from $\vect{z}$ for the Reconstruction Loss ($\loss{Recon}$). The CAR framework operates on the main network's probability maps and is driven by two key components:

\noindent\textbf{Distribution Loss ($\loss{Dist}$):} First, we construct a “reliable target" tensor, $\vect{Y}$. This composite tensor leverages the binary mask $\vect{M}$: where a valid ground truth $\vect{Y}_{gt}$ exists, $\vect{Y}$ uses $\vect{Y}_{gt}$; otherwise, it uses the network's prediction $\hat{\vect{Y}}$ with its gradients detached. This creates a complete pseudo-labeled target. Second, we apply conditional latent space regularization. The autoencoder compresses the prediction $\hat{\vect{Y}}$ and the target $\vect{Y}$ into compact latent representations, $\vect{z}_{\hat{Y}}$ and $\vect{z}_{Y}$, via the encoder $G_{\theta_1}$. To ensure conditioning on $\vect{X}$, we use a concatenation approach for all latent encodings where $\vect{z} = G_{\theta_1}(\text{Concat}(\vect{X}, \mathbf{Y}))$. This latent space is regularized by the Distribution Loss ($\loss{Dist}$), which is the core of CAR, minimizing the cosine distance between the latent representations of the prediction and the target:
\begin{equation}
    \mathcal{L}_{\text{Dist}} = \frac{1}{B} \sum_{b=1}^{B} \left( 1 - \frac{\mathbf{z}_{\hat{Y}}^{(b)} \cdot \mathbf{z}_{Y}^{(b)}}{\|\mathbf{z}_{\hat{Y}}^{(b)}\| \|\mathbf{z}_{Y}^{(b)}\|} \right)
\end{equation}

Minimizing this distance forces the model's output $\hat{\vect{Y}}$ to inhabit a latent structure that is consistent with anatomically correct organ layouts, effectively teaching it a shared anatomical prior. When the model is uncertain about an unlabeled organ, this loss encourages it to produce a prediction that is anatomically plausible given the input image $\vect{X}$ and the other, correctly-predicted organs.

\noindent\textbf{Reconstruction Loss ($\loss{Recon}$):} To ensure that the encoder $G_{\theta_1}$ learns a meaningful, non-trivial compression (rather than a trivial solution), the decoder $G_{\theta_2}$ attempts to reconstruct the full probability maps from their latent representations. We use a Mean Squared Error (MSE) loss, as the autoencoder inputs are continuous probability maps. The reconstruction targets are detached to prevent the model from learning a trivial identity function:
\begin{equation}
\begin{aligned}
\loss{Recon} = {}& \loss{MSE}(G_{\theta_2}(z_{\hat{Y}}), \hat{\vect{Y}}_{\text{detach}}) \\
&+ \loss{MSE}(G_{\theta_2}(\vect{z_{Y}}), \vect{Y}_{\text{detach}})
\end{aligned} 
\end{equation}

The final objective function combines the segmentation loss with the weighted CAR losses:
\begin{equation}
\loss{Total} = \loss{Seg} + \lambda_{\text{Dist}}\,\loss{Dist} + \lambda_{\text{Recon}}\,\loss{Recon}
\end{equation}
where $\lambda_{\text{Dist}}$ and $\lambda_{\text{Recon}}$ are balancing hyperparameters. By integrating this module, the model learns powerful contextual rules, using information from labeled organs to infer the probable location and shape of unlabeled ones.

\subsection{Implementation details}
\label{sec:implementation}
\noindent\textbf{Synthetic projection and Multi-stage randomization:} We use DiffDRR \citep{diffdrr} with the trilinear renderer, producing DRRs at $512\times512$ resolution with an effective detector pixel spacing set to the native CT in-plane voxel spacing of each case (mean $\sim 0.74$ mm, range $0.56$--$0.98$ mm across CT-RATE). This corresponds to a mean in-plane coverage of approximately 385 mm, broadly consistent with the field of view of adult chest radiography. All images are resampled to a unified $512\times512$ grid at inference to reduce scale and field-of-view mismatch. The projection geometry follows CT-RATE metadata when available; otherwise we adopt SDD as $1183\,\mathrm{mm}$ and ODD as $167\,\mathrm{mm}$ with a $\pm 10\%$ drift. The patient--detector translation is applied along the detector axis as $\Delta = \mathrm{SDD} - \mathrm{ODD}$. Volumes are resampled to the native spacings specified in the metadata, clipped to $[-1000,\,2000]$\,HU, and normalized following CT-RATE practice. We render nine view angles from $0^\circ$ to $180^\circ$ in $22.5^\circ$ steps, covering the standard Anteroposterior (AP), Posteroanterior (PA), and Lateral (LA) views, and including multiple Oblique (OB) projections. After rendering, images are min–max normalized to the $[0,255]$ range and cast to \texttt{uint8} format. Hausdorff Distance is reported in millimeters using this effective detector spacing throughout the experiments.  

To expand the variability of DRR without breaking anatomy-mask alignment, we apply Multi-stage augmentation on the 3D volumes prior to projection (attenuation statistics, low-frequency bias, mild readout-like noise, sparse artifacts) and on the 2D image after projection (exposure normalization, $\pm 2^\circ$ in-plane rotation, rare polarity perturbation). Randomizations are applied deterministically with respect to each image-mask pair so that partial labels remain co-registered across all views.

We use a concise set of parameter ranges and probabilities that prove effective and stable in ablations; quantities not listed are sampled within narrow, non-critical defaults. Structure-wise HU drift is grouped by anatomy family (cardiac, great vessels, lungs, ribs, vertebrae, shoulder girdle) but sampled from the same range across groups.

All randomization modules preserve strict mask alignment and are compatible with partial-label supervision. This dataset consists of 2{,}982 CT scans with 126{,}410 variations in different views, along with their corresponding annotations.

\vspace{0.42cm}

\noindent\textbf{CAR Training and Implementation Details:} The segmentation backbone, a U-Net with an ImageNet-pretrained ResNet-50 encoder, was adapted by modifying its first convolutional block ($\texttt{conv1}$) to accept the single-channel CXR inputs, which were preprocessed and normalized to the range $[0, 1]$. Optimization is performed using the Adam algorithm \citep{Adam} with an effective batch size of $24$ (via $4\times$ gradient accumulation) and a staged learning rate decay schedule, starting at $1\times10^{-4}$ and halved after rounds of $50$, $80$, and $128$ epochs, while stability is maintained through mixed precision and gradient clipping (global norm of $1.0$). After an initial 200-epoch warm-up, the CAR regularization module, with balancing hyperparameters set to $\lambda_{\text{Dist}}=4$ and $\lambda_{\text{Recon}}=2$, is trained for a further 200 epochs, and the entire training process utilizes five-fold cross-validation with a fixed random seed of $42$ for reproducibility. The detailed model implementation is in \textbf{\ref{app:B}}. All experiments were performed on Ubuntu~24.04.1 with an Intel i9-14900K, 64\,GB RAM, and a single NVIDIA GeForce RTX 4090. DRRs are cached in host memory to accelerate batched processing.

\input{seg_full_TS}

\subsection{Study Data Design and Evaluation Strategies}
\label{sec:strategy}
\begingroup
\noindent\textbf{Segmentation Accuracy:} We first quantify the segmentation fidelity of the 113 chest cases from the TotalSegmentator (TS) dataset \citep{TS}, which is curated with high-precision human annotations as a challenging zero-shot evaluation test set. For each test volume, DRRs are generated at the same discrete projection angles used during training; this angle-specific design is intentional, as it lets us measure how performance varies as a function of acquisition view (PA, LA, oblique), which would be obscured by averaging over uniformly random poses. The primary metric is the Dice similarity coefficient (DSC), used in conjunction with the Hausdorff Distance (HD) to quantify spatial discrepancies; both metrics are computed per-class and macro-averaged across all 54 anatomical categories.

To assess behavior on real radiographs, we randomly selected an external test set of PA and LA view radiographs from the ChestX-ray14 (CXR14) dataset \citep{CXR14}, CheXpert (CXP) dataset \citep{CheXpert}, MIMIC-CXR (MIMIC) dataset \citep{MIMIC}, and Shenzhen Tuberculosis Chest X-rays (SZ) dataset \citep{ShenzhenTB}. Specifically, we use the 10 SZ and 10 CXR14 samples for PA evaluation and the 10 CXP and 10 MIMIC samples for LA evaluation. An experienced chest physician then refined the model-generated segmentations on this 40-sample set, correcting inaccuracies to the fullest extent feasible. Because the expert started from model outputs rather than producing independent blinded annotations, these refined masks are not an unbiased ground truth; we therefore interpret the resulting DSC and HD as a measure of how efficiently the predictions can be corrected into clinically acceptable outputs rather than as an absolute estimate of segmentation accuracy.

\vspace{0.1cm}
\noindent\textbf{Comparison with External X-ray Foundation Models:} We compare two recent radiology-specialized X-ray foundation models, FluoroSAM \citep{fluorosam} and Rad-DINO \citep{raddino}, under the same DRR test set and expert-refined real-CXR set. FluoroSAM is applied directly using its native prompting protocol with no further training. Rad-DINO is plugged in as the encoder of the AnyCXR segmentation network and retrained end-to-end under the same MSDR and CAR pipeline. Because both systems differ from AnyCXR in training objective and label space, we present these as indicative comparisons rather than head-to-head benchmarks.

\vspace{0.1cm}
\noindent\textbf{Model-Agnostic Validation of CAR:} To assess whether CAR provides a backbone-independent benefit, we substitute the U-Net segmentation network with two architecturally distinct alternatives, DenseUNet, a densely-connected CNN inspired by fully convolutional DenseNets \citep{FCDenseNet}, SegFormer, a transformer-based segmenter \citep{SegFormer}, and additionally retain the Rad-DINO encoder introduced above \citep{raddino}. Each backbone is trained under the Full MSDR strategy with and without CAR, holding all other training settings identical to AnyCXR.

\vspace{0.1cm}
\noindent\textbf{Scaling Synthetic Supervision:} Our default configuration renders 126{,}410 DRRs from 2{,}982 CT volumes curated from CT-RATE under strict TotalSegmentator QC thresholds. To test how our approach scales, we construct two additional configurations that jointly expand the source CT pool and the number of rendered projections: 500{,}000 DRRs from 10{,}000 CTs, and 1{,}000{,}000 DRRs from 25{,}000 CTs. The enlarged pools are obtained by progressively relaxing the QC thresholds, which admits noisier anatomical labels in exchange for broader patient coverage and characterizes the trade-off between supervision scale and label fidelity. All scaled configurations are trained with and without CAR to test whether CAR's contribution persists once supervision is abundant.
\endgroup

\begin{figure*}[!htbp]
    \centering
    \includegraphics[width=0.80\linewidth, keepaspectratio]{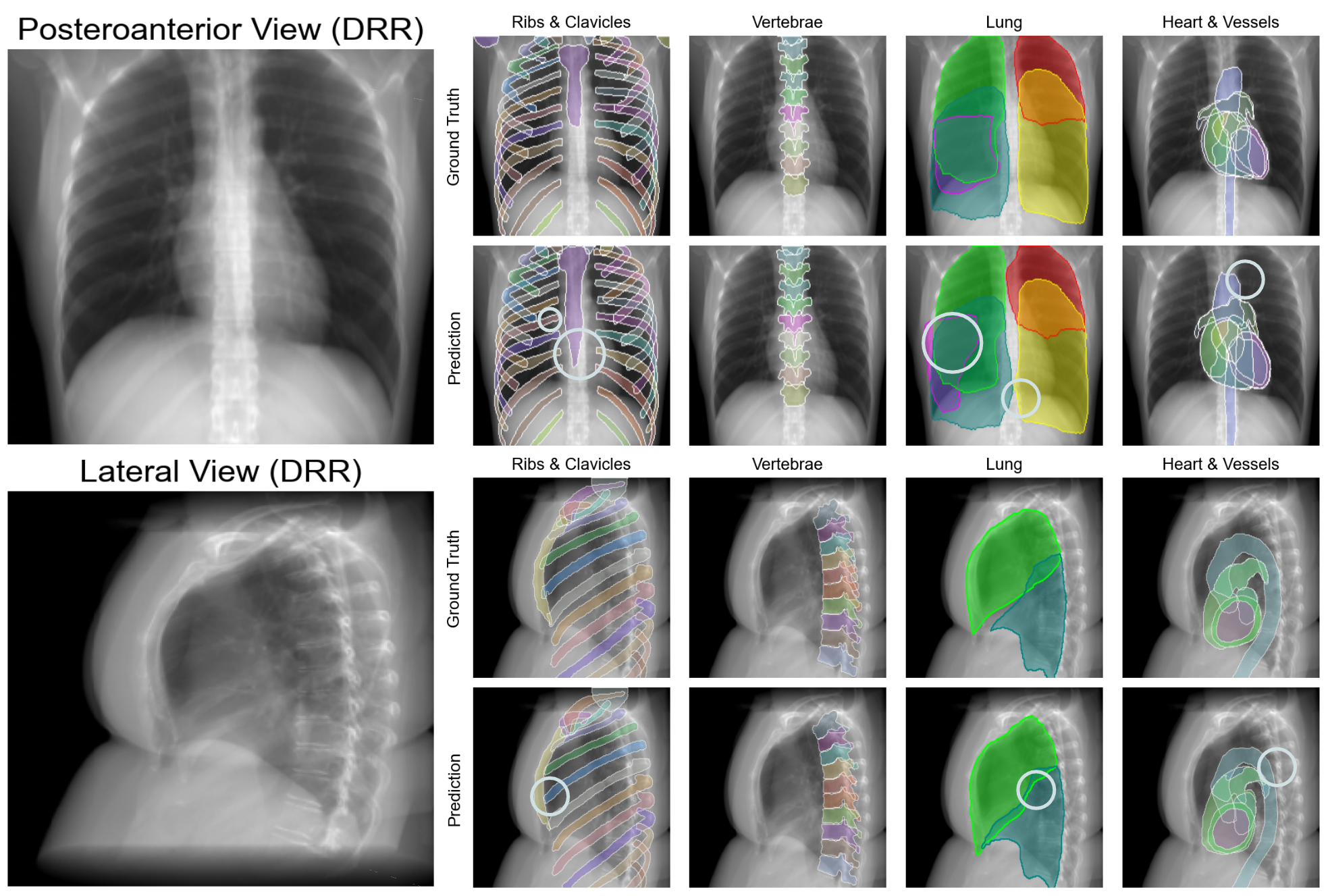}
    \caption{\small{\textbf{Segmentation Comparison on Synthetic DRR:} This figure presents a side-by-side comparison of ground truth and model Prediction for major anatomical groups on the TotalSegmentator dataset. White circles indicate specific regions where the prediction deviated from the ground truth. While overall organ localization is highly accurate, the primary challenge is focused on subtle boundary adherence and fine detail resolution. The segmentation errors, particularly the fine organ boundaries of lungs and hearts, stem from the inherent difficulty in precisely delineating edges in the 2D projection plane, where overlap is severe and the model must rely heavily on learned anatomical priors to infer invisible or faint structures.}}
    \label{fig:PALADRR}
\end{figure*}

\vspace{0.1cm}
\noindent\textbf{Downstream Clinical Applications:} To evaluate the functional value of the learned anatomy, we integrate the predicted masks into three downstream tasks that model clinical workflows, especially in an initial physical examination. These tasks serve as end-to-end validations that the masks are not only geometrically accurate but also informative for preliminary diagnosis and quantification.

\noindent\textit{(1) Cardiothoracic Ratio (CTR):}  We use the heart and lung masks from PA views to automatically derive the CTR. This is achieved by estimating the maximum cardiac and internal thoracic widths and computing the conventional $(\text{MRD} + \text{MLD}) / \text{ID}$ ratio \citep{CTR_measure}. The evaluation for this task uses 1,593 positive cases and 1,593 negative cases from the CXR14 dataset and validates the stability of predicted masks for supporting basic, size-based quantification.

\vspace{0.1cm}
\noindent\textit{(2) Thoracic Spine Curvature Assessment (SPCA):} Leveraging instance-level vertebral segmentations (T2–T12), we compute a measure of thoracic spine deviation by assessing the average displacement of vertebral centroids from a fitted centerline \citep{cobb}. This task is designed to provide a preliminary assessment of spinal curvature, stratifying cases into levels of severity to identify those that may require further evaluation or immediate attention. This task is evaluated on an in-house dataset with 12 positive cases provided by Shanxi Medical University and 2601 positive and 2601 negative cases from the MIMIC dataset. We clinically validated this methodology with a scoliosis expert, who confirmed its alignment with physical examination principles for identifying significant curvature.

\vspace{0.1cm}
\noindent\textit{(3) Disease Classification with Segmentation-Guided Inputs:} To evaluate the diagnostic utility of anatomical guidance, we compare a disease classifier's performance under two conditions using the full CXR14 dataset, which comprises 112{,}120 X-ray images with disease labels from 30{,}805 unique patients. We measured the Area Under the ROC Curve (AUC) \citep{AUC} for all diseases, comparing a model trained on raw CXRs against one trained on a multi-channel input that includes the CXR image along with the predicted lung, bone, and heart masks \citep{SegmentationImprovement}.

\input{ablation_study_dsc}

\begingroup
\subsection{Code Availability and Inference Tools}
\label{sec:code_availability}

The source code for training and inference is publicly available in the \href{https://github.com/Advanced-AI-in-Medicine-and-Physics-Lab/AnyCXR}{AnyCXR GitHub repository}. Pretrained model weights for PA, LA, and oblique radiographic views are available on \href{https://huggingface.co/agaresd/anychest}{Hugging Face}. The digitally reconstructed radiograph (DRR) dataset is publicly released through \href{https://huggingface.co/datasets/agaresd/anycxr-drr}{Hugging Face Datasets}.

For direct application to external chest radiographs, we also provide a pip-installable package through \href{https://pypi.org/project/anycxr/}{PyPI}. The command-line interface supports inference without additional training or manual annotation and generates segmentation masks for 54 anatomical structures. Supported input formats include PNG, JPEG, and DICOM, with outputs available as per-class masks and visualization overlays.

The GitHub repository provides installation instructions, example inference commands, class definitions, and evaluation scripts.

\endgroup

%% file: seg_full_TS.tex
\begingroup
\begin{table*}[!htbp]

\centering
\scriptsize
\renewcommand{\arraystretch}{1.15}
\setlength{\tabcolsep}{5pt}
\caption{Zero-shot Performance (Dice and Hausdorff Distance) by Anatomical Structure for PA/LA Views on the TotalSegmentator Dataset}
\label{tab:full_seg_TS}

\makebox[\linewidth][c]{%
\begin{tabular}{lcccc | lcccc}
\toprule
\textbf{Structure}
& \textbf{PA DSC$\uparrow$}
& \textbf{PA HD$\downarrow$}
& \textbf{LA DSC$\uparrow$}
& \textbf{LA HD$\downarrow$}
& \textbf{Structure}
& \textbf{PA DSC$\uparrow$}
& \textbf{PA HD$\downarrow$}
& \textbf{LA DSC$\uparrow$}
& \textbf{LA HD$\downarrow$} \\
\midrule
Vertebra T2 & 0.923 & 4.5 & 0.905 & 6.1 & Whole Heart & 0.929 & 14.5 & 0.935 & 11.9 \\
Vertebra T3 & 0.935 & 4.2 & 0.919 & 5.1 & Heart Atrium Left & 0.905 & 7.8 & 0.888 & 8.0 \\
Vertebra T4 & 0.930 & 5.8 & 0.907 & 7.4 & Heart Atrium Right & 0.920 & 7.6 & 0.894 & 11.5 \\
Vertebra T5 & 0.928 & 6.1 & 0.923 & 8.6 & Heart Myocardium & 0.940 & 5.6 & 0.923 & 20.5 \\
Vertebra T6 & 0.918 & 6.2 & 0.918 & 7.3 & Heart Ventricle Left & 0.936 & 5.7 & 0.868 & 40.6 \\
Vertebra T7 & 0.920 & 6.4 & 0.924 & 5.4 & Heart Ventricle Right & 0.905 & 9.5 & 0.910 & 30.0 \\
Vertebra T8 & 0.922 & 5.3 & 0.924 & 4.7 & Pulmonary Artery & 0.901 & 7.9 & 0.899 & 7.9 \\
Vertebra T9 & 0.927 & 4.8 & 0.942 & 4.4 & Aorta & 0.914 & 12.6 & 0.896 & 15.4 \\
Vertebra T10 & 0.938 & 5.6 & 0.932 & 4.1 & Lung & 0.930 & 17.9 & 0.916 & 23.2 \\
Vertebra T11 & 0.944 & 5.7 & 0.929 & 6.2 & Lung Upper Lobe Left & 0.950 & 16.7 & 0.944 & 16.6 \\
Vertebra T12 & 0.943 & 5.3 & 0.934 & 4.9 & Lung Lower Lobe Left & 0.944 & 17.9 & 0.934 & 19.6 \\
Clavicle Left & 0.946 & 3.9 & 0.914 & 13.7 & Lung Upper Lobe Right & 0.941 & 18.0 & 0.921 & 29.0 \\
Clavicle Right & 0.936 & 5.3 & 0.922 & 5.7 & Lung Middle Lobe Right & 0.872 & 20.7 & 0.853 & 28.9 \\
Humerus Left & 0.951 & 2.6 & 0.841 & 8.6 & Lung Lower Lobe Right & 0.943 & 16.2 & 0.928 & 22.0 \\
Humerus Right & 0.949 & 2.7 & 0.847 & 8.8 & Sternum & 0.918 & 12.8 & 0.923 & 20.2 \\
Rib Left 1 & 0.933 & 5.9 & 0.854 & 8.0 & Rib Right 1 & 0.938 & 5.0 & 0.884 & 6.3 \\
Rib Left 2 & 0.898 & 13.6 & 0.866 & 5.3 & Rib Right 2 & 0.906 & 8.6 & 0.861 & 6.9 \\
Rib Left 3 & 0.921 & 6.7 & 0.897 & 5.0 & Rib Right 3 & 0.910 & 7.6 & 0.877 & 5.6 \\
Rib Left 4 & 0.925 & 8.3 & 0.891 & 8.0 & Rib Right 4 & 0.910 & 8.9 & 0.886 & 13.7 \\
Rib Left 5 & 0.918 & 7.8 & 0.876 & 7.6 & Rib Right 5 & 0.921 & 8.4 & 0.856 & 13.9 \\
Rib Left 6 & 0.924 & 7.4 & 0.850 & 16.1 & Rib Right 6 & 0.929 & 6.6 & 0.822 & 9.8 \\
Rib Left 7 & 0.921 & 6.2 & 0.854 & 9.7 & Rib Right 7 & 0.931 & 7.1 & 0.874 & 13.4 \\
Rib Left 8 & 0.914 & 6.2 & 0.742 & 74.7 & Rib Right 8 & 0.927 & 6.2 & 0.843 & 15.5 \\
Rib Left 9 & 0.906 & 6.6 & 0.773 & 63.8 & Rib Right 9 & 0.925 & 5.7 & 0.848 & 34.8 \\
Rib Left 10 & 0.921 & 4.8 & 0.828 & 36.1 & Rib Right 10 & 0.937 & 5.4 & 0.865 & 27.0 \\
Rib Left 11 & 0.915 & 4.4 & 0.904 & 5.9 & Rib Right 11 & 0.940 & 3.1 & 0.898 & 6.2 \\
Rib Left 12 & 0.881 & 12.8 & 0.882 & 3.8 & Rib Right 12 & 0.899 & 5.9 & 0.857 & 4.0 \\
\bottomrule
\end{tabular}
}% end makebox

\end{table*}
\endgroup

%% file: ablation_study_dsc.tex
\begingroup
\begin{table*}[htbp!]

\centering
\footnotesize
\caption{Ablation study of DSC for major views and organ sub-groups}
\label{tab:abalation_DSC}
\newsavebox{\ablationDSCBox}
\setlength{\tabcolsep}{4pt}
\renewcommand{\arraystretch}{1.15}
\sbox{\ablationDSCBox}{%
\begin{tabular}{llcccccc}
\toprule
\textbf{View} & \textbf{Method} & \textbf{Overall$\uparrow$} & \textbf{Heart \& Vessels$\uparrow$} & \textbf{Lung$\uparrow$} & \textbf{Rib$\uparrow$} & \textbf{Vertebrae$\uparrow$} & \textbf{Other Bone$\uparrow$} \\
\midrule
PA & Plain 
& $90.86 \pm 8.15$
& $90.84 \pm 4.54$
& $93.47 \pm 4.52$
& $90.07 \pm 12.13$
& $90.21 \pm 11.62$
& $93.53 \pm 6.58$ \\
PA & MSDR V1 
& $93.75 \pm 3.61^{\dagger}$
& $93.05 \pm 2.85^{\circ}$
& $94.27 \pm 2.57$
& $\mathbf{93.38 \pm 5.03}^{\dagger}$
& $94.15 \pm 1.90^{\ddagger}$
& $95.39 \pm 2.98$ \\
PA & MSDR V2 
& $94.03 \pm 4.86^{\dagger}$
& $\mathbf{93.92 \pm 3.73}^{\circ}$
& $\mathbf{95.57 \pm 2.73}$
& $93.08 \pm 6.66^{\dagger}$
& $94.64 \pm 3.22^{\ddagger}$
& $\mathbf{95.85 \pm 3.97}$ \\
PA & MSDR Full 
& $\mathbf{94.13 \pm 4.02}^{\dagger}$
& $93.30 \pm 3.62^{\circ}$
& $95.30 \pm 3.88$
& $93.65 \pm 7.34^{\dagger}$
& $\mathbf{94.67 \pm 2.85}^{\ddagger}$
& $95.73 \pm 3.98$ \\
\midrule
LA & Plain 
& $83.39 \pm 8.15$
& $91.27 \pm 3.91$
& $90.30 \pm 3.80$
& $76.44 \pm 13.24$
& $89.39 \pm 6.82$
& $82.49 \pm 18.72$ \\
LA & MSDR V1 
& $86.94 \pm 5.66^{\dagger}$
& $\mathbf{92.27 \pm 2.62}^{*}$
& $92.81 \pm 1.87$
& $80.94 \pm 9.14^{\ddagger}$
& $\mathbf{93.14 \pm 1.48}^{\ddagger}$
& $86.65 \pm 7.38$ \\
LA & MSDR V2 
& $87.60 \pm 7.45^{\dagger}$
& $91.82 \pm 3.36$
& $\mathbf{93.89 \pm 1.29}$
& $82.50 \pm 11.55^{\dagger}$
& $93.03 \pm 2.72^{\ddagger}$
& $86.19 \pm 11.69$ \\
LA & MSDR Full 
& $\mathbf{89.90 \pm 4.73}^{\dagger}$
& $92.23 \pm 3.79$
& $\mathbf{93.89 \pm 2.37}$
& $\mathbf{86.24 \pm 9.19}^{\dagger}$
& $93.88 \pm 1.65^{\ddagger}$
& $\mathbf{90.53 \pm 8.42}$ \\
\bottomrule
\end{tabular}%
}%
\makebox[\linewidth][c]{%
\begin{minipage}{\wd\ablationDSCBox}
\centering
\usebox{\ablationDSCBox}
\vspace{0.4em}

{\footnotesize
\textit{Notes.} 
Statistical significance was assessed against Plain using a two-sided paired Wilcoxon signed-rank test within the same row: $^{*}p < 0.05$, $^{\circ}p < 0.01$, $^{\ddagger}p < 0.001$, $^{\dagger}p < 0.0001$.
}
\end{minipage}
}%
\end{table*}
\endgroup

%% file: experiment.tex
\section{Experimental Results}
\label{sec:results}

In this section, we present a detailed analysis of our framework's performance, structured to evaluate its core components and clinical utility. First, we quantify the segmentation accuracy and generalization of the model on both synthetic and real-world benchmarks (Section~\ref{sec:accuracy_results}). Second, we conduct ablation studies to isolate and validate the contributions of the AnyCXR generation engine and the CAR framework (Section~\ref{sec:ablation_results}). Finally, we demonstrate the functional value of the learned anatomy through three distinct downstream clinical applications (Section~\ref{sec:downstream_results}).

\subsection{Segmentation Accuracy and Generalization}
\label{sec:accuracy_results}

\begingroup

We first evaluate the segmentation performance of the proposed AnyCXR-CAR model. We report the DSC and HD across all 54 anatomical classes on the zero-shot TS test set in Table~\ref{tab:full_seg_TS}, while the corresponding metrics for the oblique view are provided in detail in \textbf{\ref{app:E}}. Performance is broken down by PA and LA views to demonstrate robust prediction across geometric variations. Specifically, the model achieved a mean DSC of 92.4\% and an HD of 8.06~mm in the PA view, while maintaining strong performance in the challenging LA view with a DSC of 88.7\% and an HD of 14.95~mm.
\endgroup

\begin{figure*}[!bhtp]
    \centering
    \includegraphics[width=0.80\linewidth, keepaspectratio]{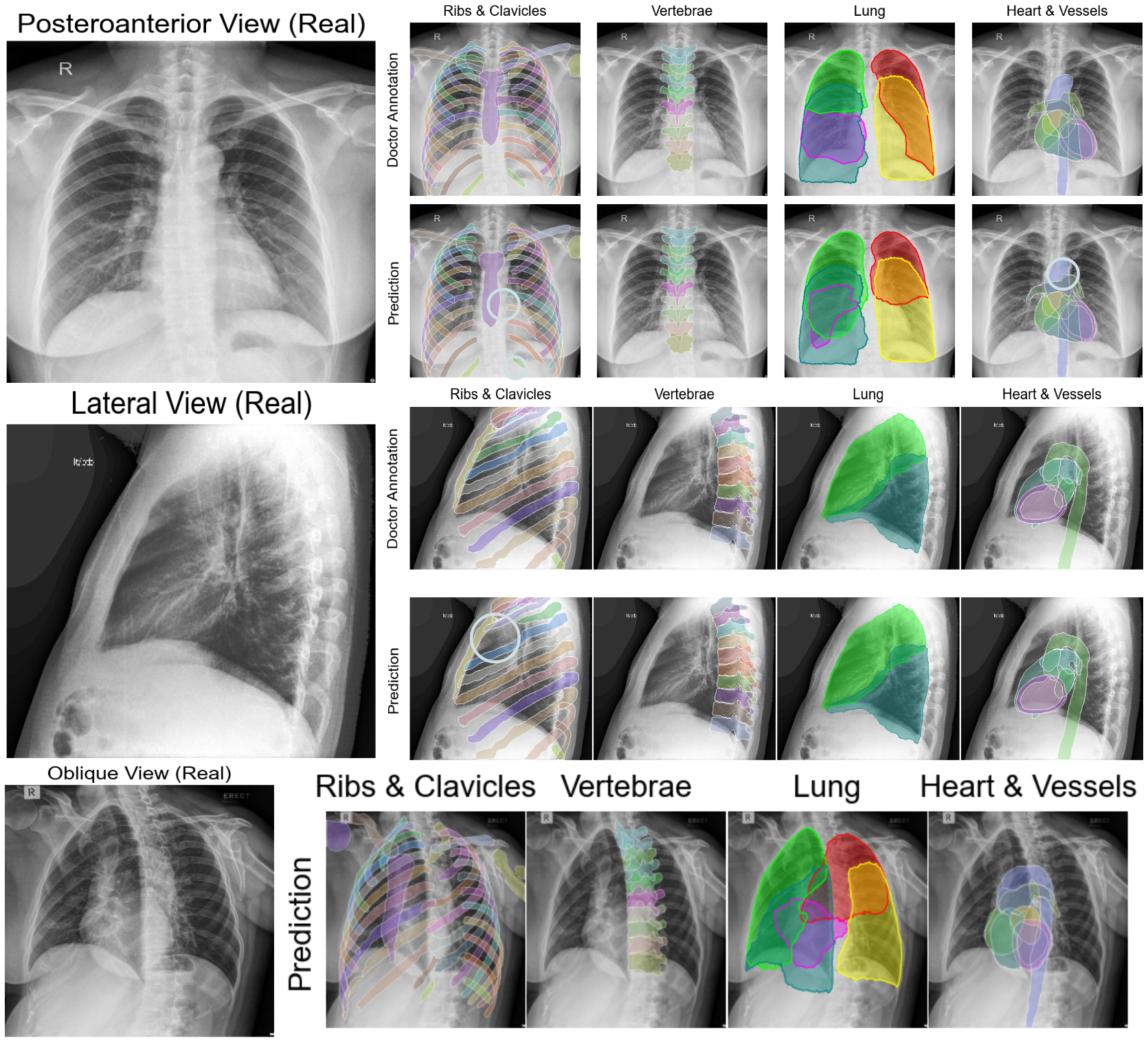}
    \caption{\small{\textbf{Detailed Comparison of Segmentation on Real-World CXRs and Expert Refinement:} This figure presents a comparison of the model's prediction against the expert corrections on real PA and LA view radiographs, illustrating the editing required to obtain clinically acceptable masks. White circles indicate specific regions where the prediction deviated from the expert annotation. While AnyCXR provides a highly effective initial segmentation proposal, these differences reflect areas of necessary expert editing. Specific challenges arise from image contrast variations and overexposure in clinical data, making detailed editing difficult, particularly for fine structures like the ribs and lungs. Furthermore, a non-trivial portion of the divergence relates to the inherent clinical ambiguity of 2D projection, such as precise lung lobe placement and boundary definition in superimposed regions, where the expert's final delineation corrects the prediction based on clinical consensus.}}
    \label{fig:real_compare}
\end{figure*}
\input{seg_Doctor}
\begingroup
Overall, the framework demonstrates good performance across all of the major organ groups, as shown in Fig.~\ref{fig:PALADRR}, but it is not perfect, as boundary adherence remains a significant challenge, especially for regions with low contrast or severe superposition. Similar boundary ambiguity issues are reflected in the oblique view DRR segmentation in \textbf{\ref{app:D}}. This issue is particularly notable at the boundaries of the Lung Group (DSC of around 91.6\% in LA view), where the segmentation boundary derived from 3D CT projection is physically more accurate but inherently more complex than traditional human-drawn 2D annotations, leading to a higher expectation of error. This boundary ambiguity similarly affects the delineation of the Sternum, individual Ribs, and the middle of the spine (Vertebrae T4-T8, DSCs around 93\% in PA view). It is also critical to note that the performance on the lower ribs (e.g., Rib Left 12, DSC 88.1\% PA view) is not consistent because the lower ribs are often partially or entirely excluded from the original 3D CT training volumes, leading to model degradation due to a lack of complete training data.
\endgroup

To assess how closely the predictions approximate expert-acceptable masks, we report the macro-averaged DSC against the expert-refined real-world set in Table~\ref{tab:doc_seg}, complemented by Fig.~\ref{fig:real_compare}. Because the expert started from model outputs, these numbers reflect the agreement between predictions and their refined versions rather than an unbiased estimate of accuracy, and are best read as an indication of how little editing was required. In the PA view, the Bone group reached a DSC of 97.8\%, while the LA view showed high agreement for soft tissues, with DSCs of 98.2\% for the Lung group and 96.7\% for the Heart and Vessels.

\textbf{External X-ray foundation models.} Following the protocol of Section~\ref{sec:strategy}, off-the-shelf FluoroSAM \citep{fluorosam} produced limited performance across all organ groups on both the DRR test set and the expert-refined CXRs (App.~\ref{app:H}), reflecting the gap between its native fluoroscopic prompt taxonomy and our 54-class chest anatomy. Rad-DINO \citep{raddino}, used as the encoder backbone within the AnyCXR pipeline, performed competitively on most organ groups (App.~\ref{app:J}) and is reported jointly with the other backbone substitutes in Section~\ref{sec:ablation_results}. As noted in Methods, both comparisons are indicative rather than head-to-head, given the different training objectives and label spaces.

\subsection{Ablation Studies and Component Analysis}
\label{sec:ablation_results}

To isolate the contributions of our key innovations, we perform ablation studies, primarily focusing on the impact on macro-averaged DSC in the held out test set.

\begin{figure*}[!htbp]
    \centering
    \centerline{\includegraphics[width=1.78\columnwidth]{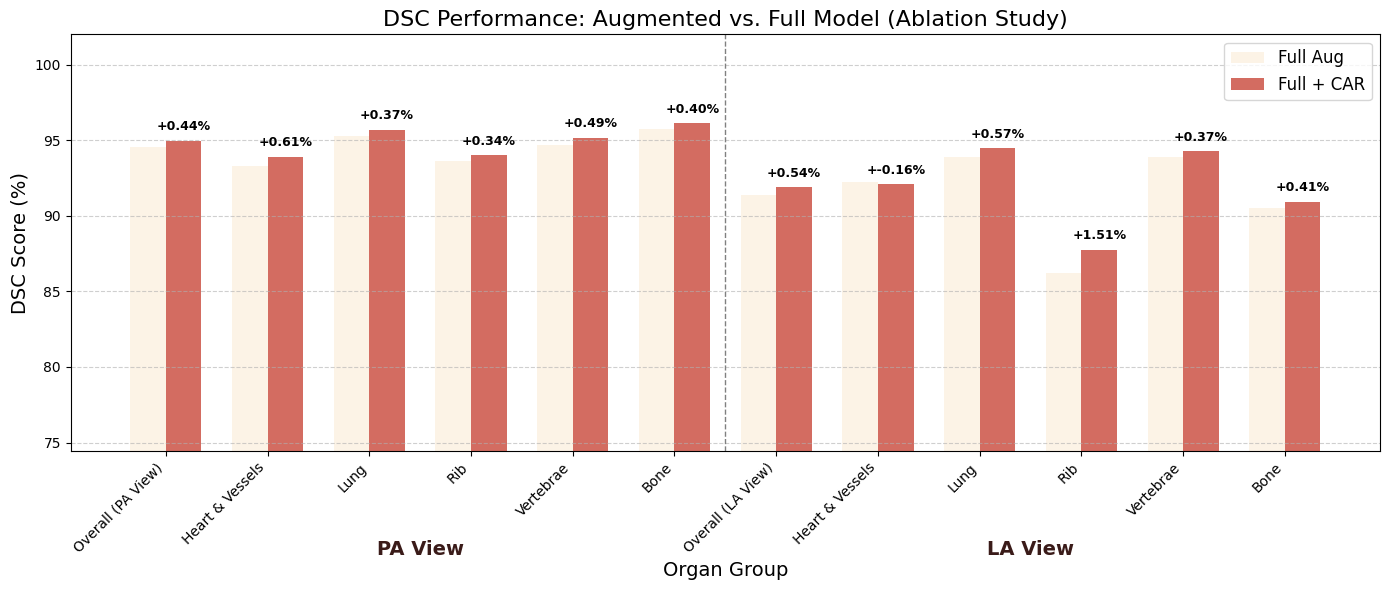}}

 \caption{\small{\textbf{Ablation Study of the CAR model:} This chart compares the DSC performance of the Full Aug model (MSDR + PostHoc augmentations) against the final CAR + Aug framework. The CAR framework consistently improves performance across all groups, with the largest gain ($+1.51\%$) seen in the challenging Lateral Ribs group, validating CAR's efficacy in enforcing anatomical priors where augmentation alone is insufficient.}}
	\label{fig:abalation}
\end{figure*}

\begingroup
 \textbf{Contribution of CAR Regularization and MSDR:} In our ablation analysis, we investigate the contribution of both the data augmentation strategy and the proposed CAR method. The five configurations studied are referred to by the following shorthand notation throughout the results section and figures: \textbf{Plain} refers to the baseline UNet architecture trained without any data augmentation; \textbf{MSDR V1} denotes the UNet model trained exclusively using post-processing augmentation techniques; \textbf{MSDR V2} (or Non-Posthoc Augmentation) represents the UNet model trained only using 3D augmentation; \textbf{MSDR Full} represents the UNet model trained with the combined set of all augmentation techniques; and finally, \textbf{Full + CAR} represents our full proposed architecture, which integrates the CAR model into the UNet framework and is trained with the optimal \textbf{MSDR Full} strategy.

The progressive DSC gains detailed in Table~\ref{tab:abalation_DSC} validate the effectiveness of our comprehensive data strategy, and the parallel HD improvements can be found in \textbf{\ref{app:F}}. In the PA view, the MSDR V2 configuration achieved an Overall DSC of 94.03\%, surpassing the MSDR V1 baseline of 93.75\%, confirming the benefit of volumetric randomization. For the more challenging Lateral (LA) view, the synergy of both strategies proved critical: the MSDR Full configuration achieved an Overall DSC of 89.90\%, outperforming the Plain baseline of 83.39\% and the MSDR V1 model at 86.94\%.

\input{three_modelablation}
The final integration of the CAR framework delivers the definitive performance gain, as visually confirmed in Fig.~\ref{fig:abalation}. The complete framework (\textbf{Full + CAR}) improves over the strong MSDR Full baseline in both views, with structure-level paired gains of +0.43\% in PA and +0.81\% in LA (Table~\ref{tab:paired_dsc_statistics}). Most notably, CAR addresses the topological ambiguity in the challenging lateral projection, driving a substantial +1.51\% DSC improvement in the Lateral Rib group and a +0.57\% boost in the Lateral Lung group. This specific enhancement confirms that CAR effectively regularizes shape constraints in highly superimposed regions where standard data augmentation plateaus.
\endgroup

\begingroup

\textbf{Comparison with External X-ray Foundation Models:} The per-group DSC comparison between AnyCXR and FluoroSAM is summarized in Fig.~\ref{fig:fsam_compare}, and the corresponding qualitative predictions on real CXRs are illustrated in Fig.~\ref{fig:Fsam_real} with detailed per-class results in \textbf{\ref{app:G}} and \textbf{\ref{app:H}}. FluoroSAM yields unstable and limited performance across organ groups in both DRR and real radiograph settings. We report the performance of Rad-DINO in Table~\ref{tab:three_model_improve}, with detailed per-class results in \textbf{\ref{app:I}} and \textbf{\ref{app:J}}. Rad-DINO, used as the AnyCXR encoder with CAR enabled, reaches an overall DSC of 86.3\% in PA and 81.3\% in LA. Given its substantially smaller backbone capacity, this comparison is indicative rather than head-to-head.

\textbf{Model-Agnostic Validation of CAR:} We report DenseUNet, SegFormer, and Rad-DINO trained with and without CAR under the Full MSDR strategy. As shown in Table~\ref{tab:three_model_improve}, CAR improves the overall DSC across all three backbones and both views, although a few organ-level categories show small decreases or no change. In the PA view, the overall DSC increases by +1.9\% for DenseUNet, +1.3\% for SegFormer, and +0.6\% for Rad-DINO. In the more challenging LA view, CAR also improves the overall DSC for all three backbones, with particularly consistent gains in Vertebrae. These results confirm that CAR functions as a general-purpose shape regularizer that complements both CNN and transformer encoders.

\begin{figure*}[!htbp]
    \centering
    \centerline{\includegraphics[width=1.78\columnwidth]{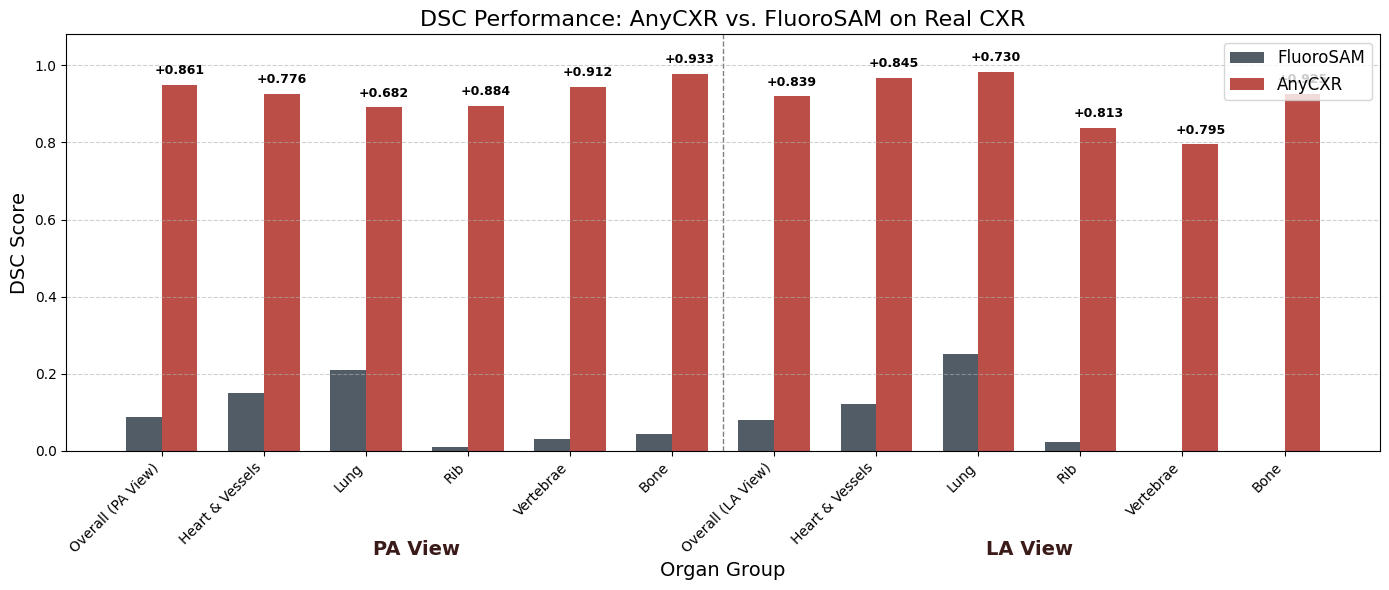}}
    
 \caption{\small{\textbf{Real CXR Segmentation Performance Comparison:} This chart compares the DSC performance of FluoroSAM against AnyCXR on expert-refined real CXR annotations across PA and LA views. AnyCXR achieves substantially higher DSC scores across all evaluated organ groups, demonstrating stronger generalization to real CXR segmentation.}}
	\label{fig:fsam_compare}
\end{figure*}

\begin{figure*}[!htbp]
    \centering
    \includegraphics[width=0.7\textwidth]{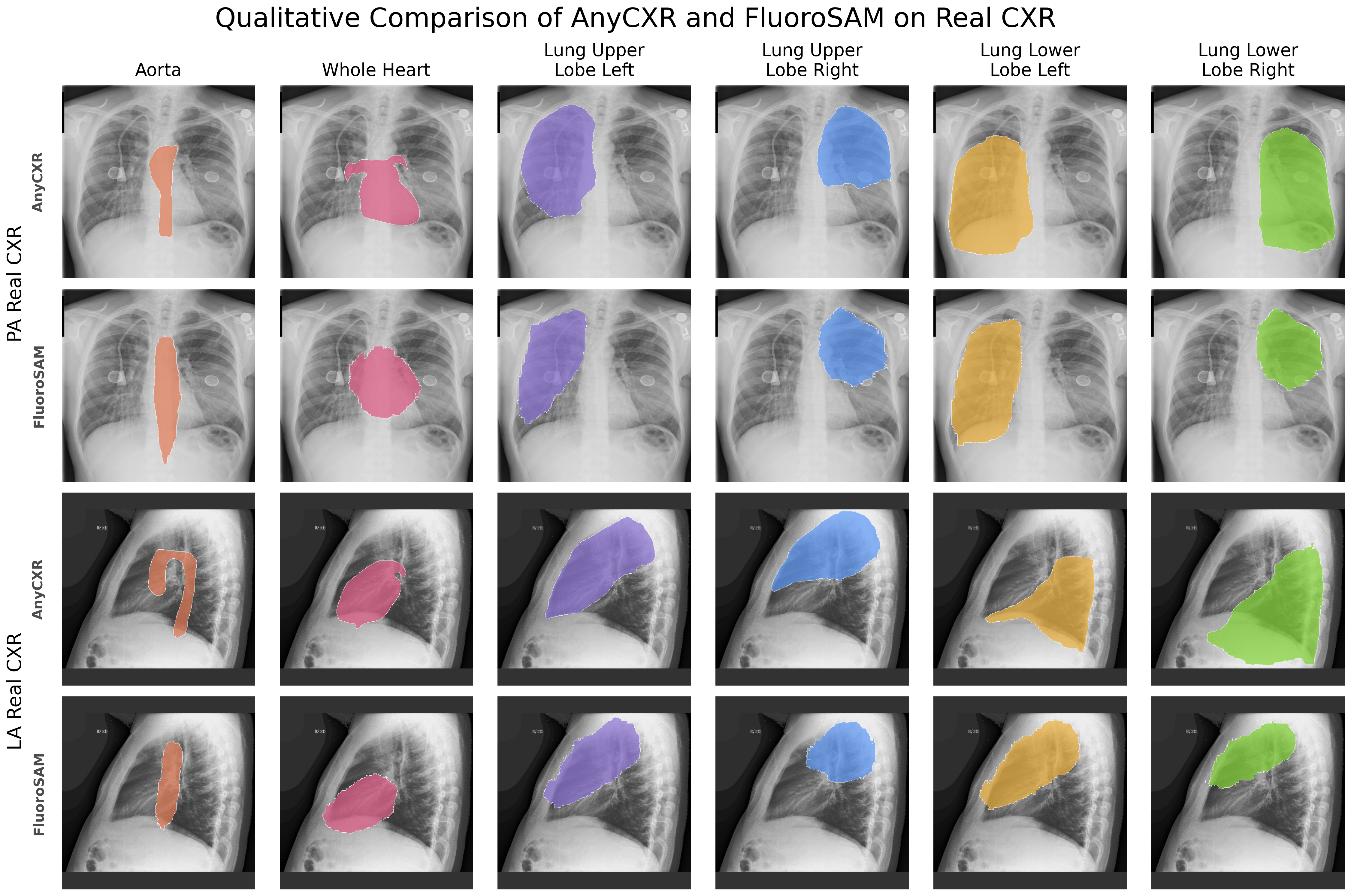}
    \captionsetup{justification=centering, singlelinecheck=false}

     \caption{Segmentation comparison of FluoroSAM and AnyCXR on CXR images in PA and LA views.}

    \label{fig:Fsam_real}
\end{figure*}

\textbf{Scaling synthetic supervision.} As shown in Figure~\ref{fig:scaling}, expanding the pool under relaxed QC yields only marginal average gains of 1.07\% Dice, while CAR continues to deliver consistent improvements across all scales. These results confirm that label fidelity, not projection volume, is the binding constraint on synthetic supervision, and that CAR's regularization remains beneficial as supervision scales.
\begin{figure}[htbp]
    \centering
    \centerline{\includegraphics[width=0.95\columnwidth]{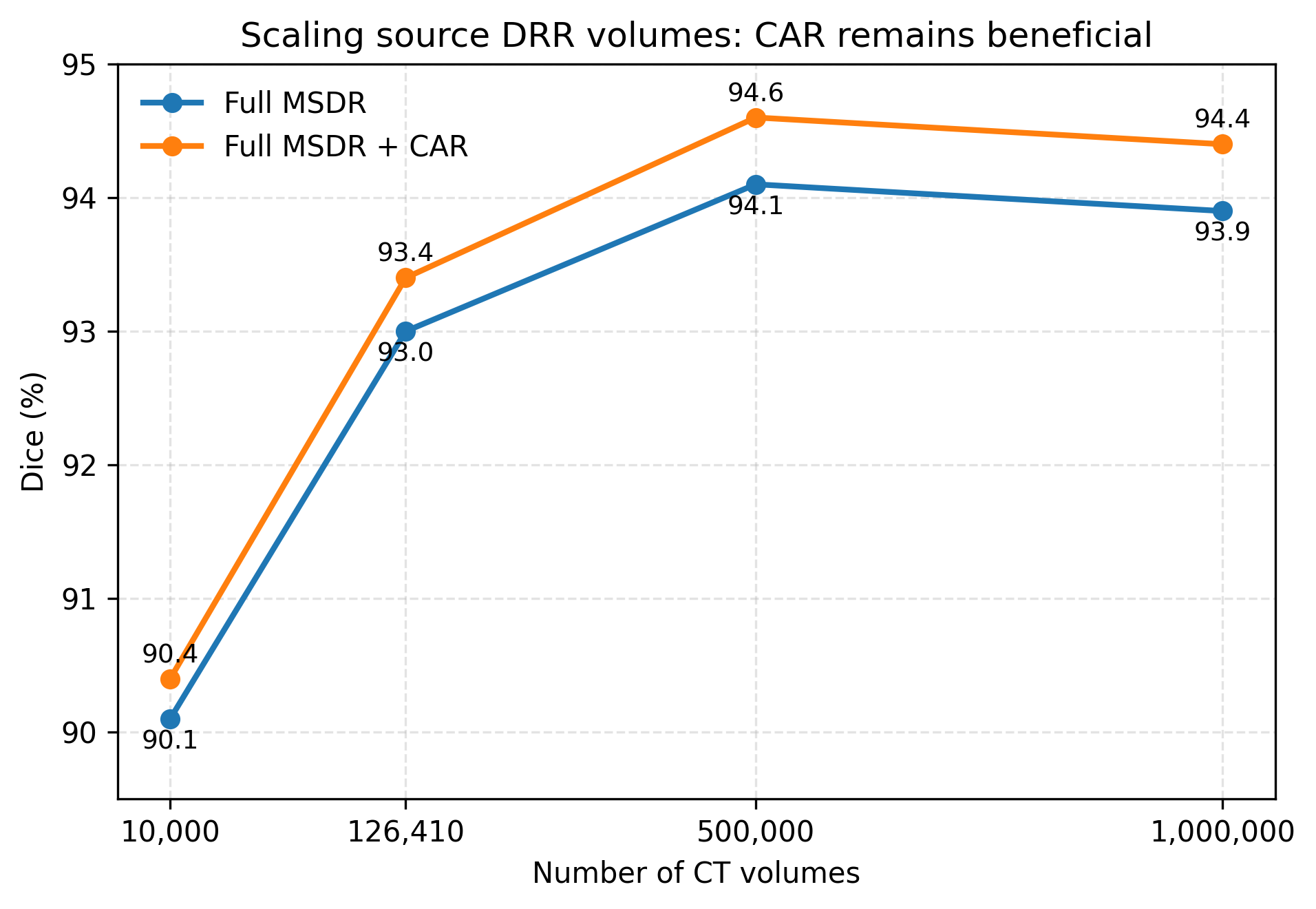}}
    
 \caption{\small{\textbf{Scaling Synthetic Supervision:} This figure demonstrates the DSC of Full MSDR with and without CAR across different configurations. The 500{,}000 and 1{,}000{,}000 DRR configurations are generated under relaxed QC thresholds.}}
	\label{fig:scaling}
\end{figure}
\endgroup

\begin{figure*}[!htbp]
    \centering
    \centerline{\includegraphics[width=1.45\columnwidth]{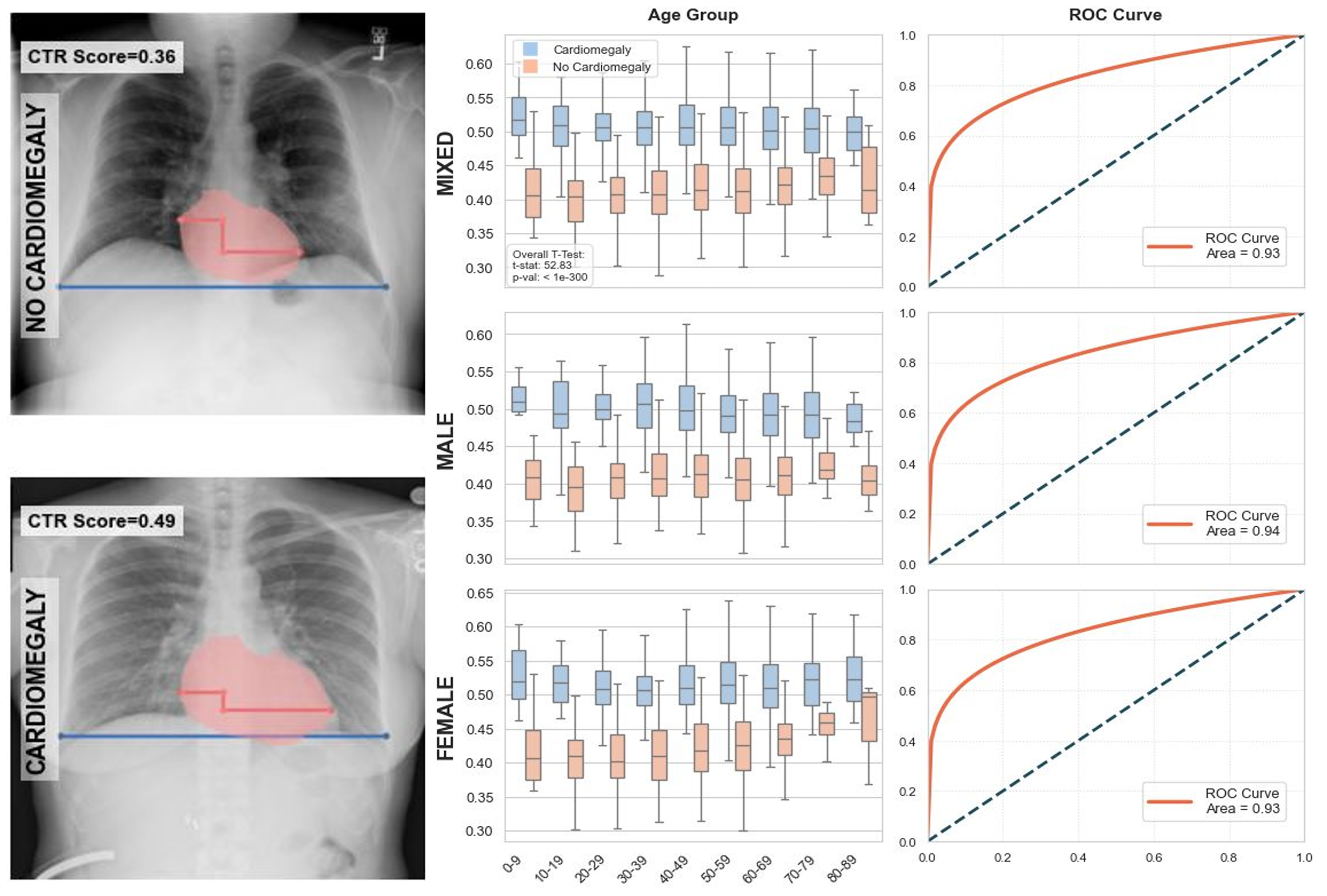}}
    
 \caption{\small{\textbf{Visualization for CTR disease detection:} This figure demonstrates the utility of the segmentation model for automatic Cardiothoracic Ratio (CTR) assessment in the PA view. The left panel shows the delineated heart and the measurement lines used to calculate the CTR score, distinguishing between a normal case and a pathological case. The right panels illustrate the ROC curves and distribution of derived CTR scores, achieving an overall AUC of 0.93 with statistically significant separation between normal and cardiomegaly subjects across age and gender groups.}}
	\label{fig:CTR}
\end{figure*}

\begin{figure*}[!htbp]
    \centering
    \centerline{\includegraphics[width=1.45\columnwidth]{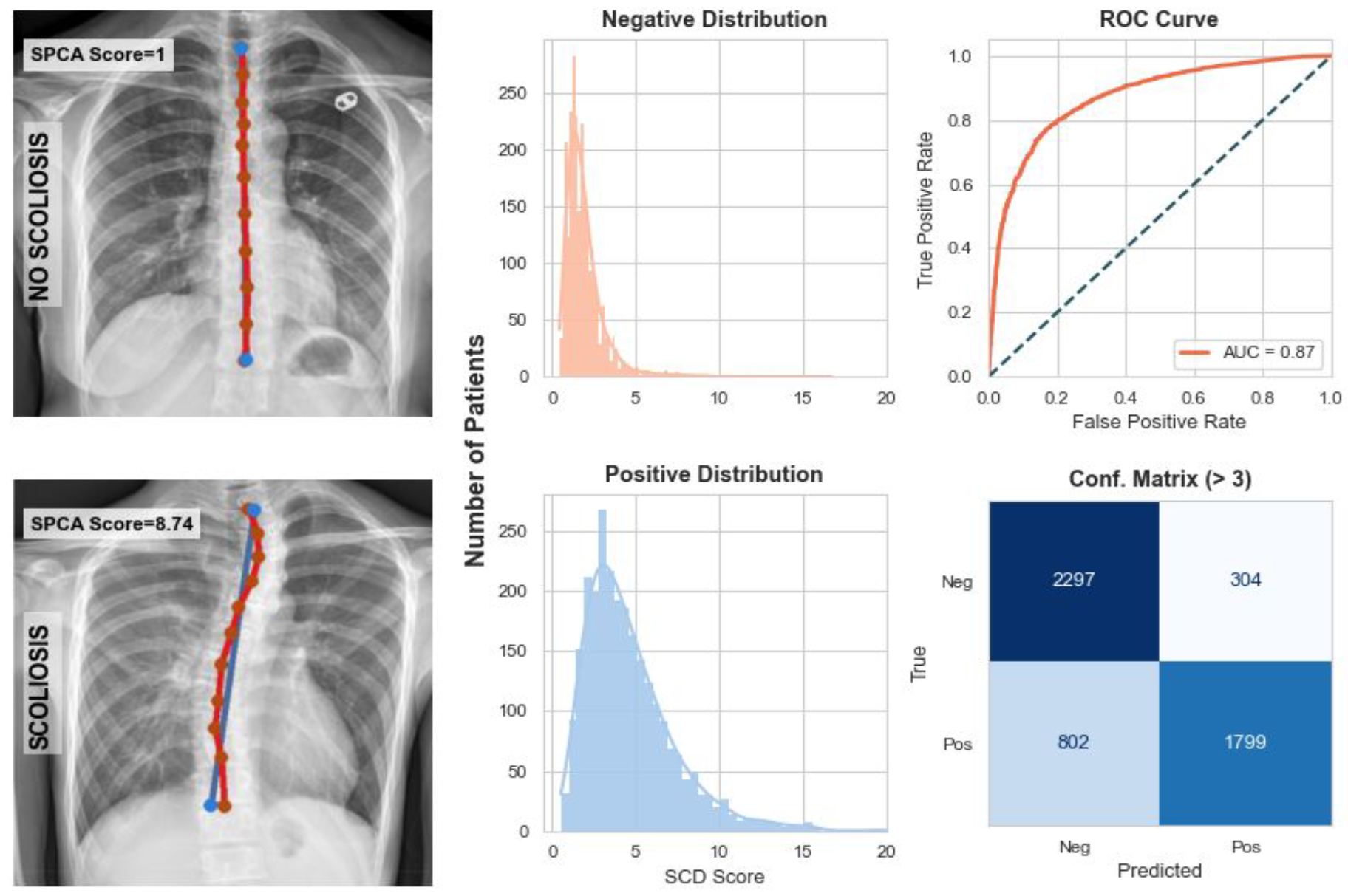}}
    
 \caption{\small{\textbf{Visualization for SPCA disease detection:} This figure demonstrates the clinical utility of instance-level vertebral segmentation for preliminary Spine Curvature Assessment (SPCA). The left panel illustrates the identification and tracking of thoracic vertebral centroids to measure deviation, distinguishing between a normal spine and a case with probable scoliosis. The center and right panels show the distribution of SPCA scores and the ROC curve, achieving an AUC of 0.87, which validates its effectiveness as a screening tool for spinal deformities.}}
	\label{fig:SPCA}
\end{figure*} 

\subsection{Downstream Clinical Utility}
\label{sec:downstream_results}

We demonstrate the functional value of our segmentations by visualizing their utility across two proposed downstream clinical tasks and quantifying their performance in disease classification.

\textbf{Cardiothoracic Ratio (CTR) Demonstration:} We demonstrate the utility of our cardiac segmentation by visually illustrating the CTR measurement in Fig.~\ref{fig:CTR}. The figure shows the accurate delineation of the heart and the measurement lines used to derive the ratio between the maximal transverse cardiac diameter and the maximal transverse thoracic diameter. Quantitatively, the derived CTR scores achieved an AUC of 0.93 with a statistically significant separation ($p \ll 0.001$) between normal and cardiomegaly subjects.

\textbf{Spine Curvature Assessment (SPCA) Validation:} We validate the utility of our instance-level vertebral segmentation by visualizing its application to SPCA in Fig.~\ref{fig:SPCA}. The accompanying figure demonstrates the model's ability to precisely identify and track the thoracic vertebral centroids. This measure allows for a preliminary, automated stratification of cases into one of three urgency levels, indicating the need for further clinical evaluation or immediate attention. In our evaluation, this metric achieved an AUC of 0.87 with a statistically significant separation ($p \ll 0.001$), validating its effectiveness as a reliable screening tool for spinal deformities.

\textbf{Segmentation-Guided Disease Classification:} We report the AUC for the detection of all 14 diseases on the CXR14 dataset in Table~\ref{tab:disease}. We compare the performance of the baseline raw CXR classifier against the Segmentation-Guided classifier. Quantitatively, the segmentation-guided model achieved a Mean AUROC of 82.30\%, representing a 2.04\% absolute increase over the 80.26\% baseline. The most significant improvements were observed in Atelectasis (+3.36\% AUC) and Pleural Thickening (+3.30\% AUC), validating the role of explicit anatomical guidance in boosting diagnostic performance.

\input{disease_detect}

%% file: seg_Doctor.tex
\begingroup
\begin{table}[!htbp]

\centering
\scriptsize
\renewcommand{\arraystretch}{1.20}
\setlength{\tabcolsep}{6pt}
\caption{Segmentation Statistics by Organ Group Following Expert Revision (PA and LA Views)}
\label{tab:doc_seg}

\makebox[\linewidth][c]{%
\begin{tabular}{lcccc}
\toprule
\textbf{Group: PA} 
& \textbf{DSC$\uparrow$}
& \textbf{Std DSC}
& \textbf{HD$\downarrow$}
& \textbf{Std HD} \\
\midrule

Heart \& Vessels
& 0.926 & 0.068 & 32.9 & 17.7 \\

Lung Group
& 0.891 & 0.058 & 61.4 & 40.4 \\

Rib Group
& 0.894 & 0.143 & 35.7 & 17.9 \\

Vertebrae Group
& 0.943 & 0.154 & 6.1 & 14.1 \\

Other Bone Group
& 0.978 & 0.046 & 5.8 & 11.6 \\

\midrule

\textbf{Group: LA}
& \textbf{DSC$\uparrow$}
& \textbf{Std DSC}
& \textbf{HD$\downarrow$}
& \textbf{Std HD} \\
\midrule

Heart \& Vessels
& 0.967 & 0.047 & 35.7 & 58.2 \\

Lung Group
& 0.982 & 0.034 & 41.9 & 74.3 \\

Rib Group
& 0.837 & 0.190 & 88.9 & 133.4 \\

Vertebrae Group
& 0.795 & 0.299 & 58.9 & 85.1 \\

Other Bone Group
& 0.925 & 0.173 & 17.9 & 45.7 \\

\bottomrule
\end{tabular}
}% end makebox

\end{table}
\endgroup

%% file: three_modelablation.tex
\begingroup
\begin{table*}[htbp]

\centering
\footnotesize
\caption{DenseUNet, SegFormer, and Rad-DINO performance of DSC with and without CAR}
\label{tab:three_model_improve}
\setlength{\tabcolsep}{4pt}
\renewcommand{\arraystretch}{1.15}

\begin{tabular}{lllcccccc}
\toprule
\textbf{View} & \textbf{Method} & \textbf{CAR} & \textbf{Overall$\uparrow$} & \textbf{Heart \& Vessels$\uparrow$} & \textbf{Lung$\uparrow$} & \textbf{Rib$\uparrow$} & \textbf{Vertebrae$\uparrow$} & \textbf{Other Bone$\uparrow$} \\
\midrule
PA & DenseUNet & Off
& 0.825
& 0.861
& 0.890
& 0.752
& 0.795
& 0.829 \\
PA & DenseUNet & On
& \textbf{0.844}
& \textbf{0.872}
& \textbf{0.900}
& \textbf{0.783}
& \textbf{0.825}
& \textbf{0.843} \\
PA & SegFormer & Off
& 0.841
& 0.875
& 0.916
& 0.742
& 0.821
& 0.852 \\
PA & SegFormer & On
& \textbf{0.854}
& \textbf{0.882}
& \textbf{0.921}
& \textbf{0.768}
& \textbf{0.836}
& \textbf{0.865} \\
PA & Rad-DINO & Off
& 0.857
& \textbf{0.897}
& 0.930
& 0.742
& 0.862
& 0.854 \\
PA & Rad-DINO & On
& \textbf{0.863}
& 0.892
& \textbf{0.934}
& \textbf{0.747}
& \textbf{0.874}
& \textbf{0.869} \\
\midrule
LA & DenseUNet & Off
& 0.712
& 0.863
& 0.853
& 0.432
& 0.783
& \textbf{0.628} \\
LA & DenseUNet & On
& \textbf{0.720}
& \textbf{0.867}
& \textbf{0.860}
& \textbf{0.442}
& \textbf{0.815}
& 0.616 \\
LA & SegFormer & Off
& 0.745
& 0.872
& 0.901
& \textbf{0.500}
& 0.752
& 0.698 \\
LA & SegFormer & On
& \textbf{0.751}
& \textbf{0.886}
& \textbf{0.904}
& 0.495
& \textbf{0.760}
& \textbf{0.709} \\
LA & Rad-DINO & Off
& 0.806
& \textbf{0.893}
& 0.911
& \textbf{0.623}
& 0.847
& 0.758 \\
LA & Rad-DINO & On
& \textbf{0.813}
& 0.885
& \textbf{0.926}
& 0.623
& \textbf{0.865}
& \textbf{0.765} \\
\bottomrule
\end{tabular}%

\end{table*}
\endgroup

%% file: disease_detect.tex
\begin{table}[thbp]
\centering
\caption{Segmentation-Guided Chest X-ray Disease Classification}
\label{tab:disease}
\resizebox{\linewidth}{!}{
\begin{tabular}{lcccc}
\toprule
\textbf{Pathology} & \textbf{Raw CXR} & \textbf{CXR w/mask} & \textbf{\(\Delta\) AUC} & \textbf{Improve \%} \\
\midrule
Mean AUROC & $80.26\%$ & $82.30\%$ & $2.04\%$ & $2.54\%$ \\
\midrule
Pleural Thickening & $76.83\%$ & $80.13\%$ & $3.30\%$ & $4.30\%$ \\
Atelectasis & $81.09\%$ & $84.45\%$ & $3.36\%$ & $4.14\%$ \\
Mass & $86.46\%$ & $89.56\%$ & $3.10\%$ & $3.59\%$ \\
Cardiomegaly & $90.14\%$ & $93.33\%$ & $3.19\%$ & $3.54\%$ \\
Effusion & $87.45\%$ & $90.39\%$ & $2.94\%$ & $3.36\%$ \\
Emphysema & $86.66\%$ & $89.38\%$ & $2.72\%$ & $3.14\%$ \\
Pneumothorax & $86.83\%$ & $89.30\%$ & $2.47\%$ & $2.84\%$ \\
Fibrosis & $77.72\%$ & $79.82\%$ & $2.10\%$ & $2.70\%$ \\
Pneumonia & $67.94\%$ & $69.48\%$ & $1.54\%$ & $2.27\%$ \\
Consolidation & $77.28\%$ & $78.69\%$ & $1.41\%$ & $1.82\%$ \\
Infiltration & $63.80\%$ & $64.71\%$ & $0.91\%$ & $1.43\%$ \\
Edema & $79.37\%$ & $80.42\%$ & $1.05\%$ & $1.32\%$ \\
Hernia & $85.36\%$ & $85.75\%$ & $0.39\%$ & $0.46\%$ \\
Nodule & $76.64\%$ & $76.72\%$ & $0.08\%$ & $0.10\%$ \\
\bottomrule
\end{tabular}
}
\end{table}

%% file: discussion.tex
\section{Discussion}

In this work, we introduced AnyCXR, a unified framework that combines Multi-stage Domain Randomization (MSDR) with Conditional Joint Annotation Regularization (CAR) to achieve robust, generalizable anatomy segmentation across clinically relevant CXR projection views using only synthetic supervision. Our experiments demonstrate that the model effectively bridges the domain gap between synthetic and clinical radiographs. In PA views, AnyCXR achieves precise boundary adherence, while in lateral views it maintains strong volumetric recovery despite substantial anatomical overlap that typically complicates segmentation. These findings indicate that the MSDR engine successfully captures the breadth of visual and geometric variability encountered in real world imaging. Clinical benchmarking further confirms the practical utility of the framework. As shown in Table~\ref{tab:doc_seg}, AnyCXR performs reliably as a pre-annotation tool and yields high agreement with expert labels across multiple anatomical structures. Ablation experiments in Table~\ref{tab:abalation_DSC} highlight the role of CAR in improving topological consistency, particularly in regions where structures overlap or exhibit subtle boundaries. Beyond segmentation accuracy, AnyCXR also demonstrates value for downstream clinical tasks. The anatomically informed features support population level screening workflows, achieving high accuracy for cardiothoracic ratio estimation and spine curvature assessment, and they provide explicit spatial priors that significantly enhance disease classification performance (Table~\ref{tab:disease}). Taken together, these results suggest that AnyCXR offers a scalable and effective foundation for generalizable CXR analysis.

Although the framework demonstrates strong performance, several limitations remain, many of which stem from the synthetic data generation process. One limitation arises from the automated 3D segmentation masks used as ground truth labels for CT volumes. These masks occasionally contain voxel level jaggedness or surface irregularities that become amplified when projected into 2D. Interestingly, the model, guided by the CAR regularizer, often produces smoother and more anatomically consistent boundaries than the synthetic labels themselves. This behavior is desirable from a clinical perspective, yet it can penalize overlap based metrics on the synthetic test set and lead to lower DSC and higher HD values even when the predictions qualitatively appear superior. A second limitation relates to the divergence between anatomically correct projections derived from CT and the stylistic conventions of clinical 2D annotation. Radiologists typically delineate structures according to visible intensity transitions rather than their true anatomical extent. As a result, the model often predicts anatomically accurate boundaries that differ from the consensus appearance of standard radiographs, particularly for lung lobes and costovertebral junctions. Because AnyCXR is trained exclusively on DRR data, expert annotators frequently corrected these regions to align with clinical practice, which produced a measurable drop in performance when compared with human labels. For applications that prioritize conformity to 2D clinical conventions, fine tuning the model with a small set of expert annotated CXRs may help reconcile this semantic gap.

\begingroup
A further limitation concerns the evaluation on real CXRs. Since the reference masks were obtained by expert refinement of model predictions rather than independent blinded annotation, the resulting Dice scores may be influenced by anchoring bias. The reliability of these reference masks also varies across anatomical classes and acquisition views. Structures with relatively identifiable radiographic boundaries, such as lung-field contours, clavicles, proximal humeri, and unobscured portions of ribs and vertebrae, can generally be evaluated with greater confidence. However, anatomical superposition and projection geometry may still introduce ambiguity. In contrast, individual lung lobes, cardiac chambers, myocardium, and portions of the great vessels cannot be reliably delineated in full from conventional 2D radiographs alone. Their annotations therefore depend substantially on anatomical inference. Consequently, Dice scores calculated against expert-refined reference masks reflect agreement with expert-defined anatomical estimates rather than independent verification of the underlying projected 3D anatomy, particularly for these less observable classes.

Our downstream evaluations, including cardiothoracic ratio estimation, spine curvature assessment, and segmentation-guided classification, provide additional evidence of functional utility. Nevertheless, these results do not independently establish segmentation accuracy for all 54 anatomical classes.

Patient-specific CT-to-X-ray registration could provide a complementary evaluation reference by projecting 3D anatomical annotations onto corresponding radiographs. However, this approach requires paired imaging and accurate registration. Differences in patient positioning, respiratory state, and acquisition geometry may introduce additional uncertainty. Future evaluations could combine registered CT-derived projections with blinded multi-reader annotations of reliably visible structures to provide stronger independent validation.

A separate limitation concerns the range of acquisition geometries covered by AnyCXR. The framework targets standard PA/AP views, lateral views, and moderate oblique or positioning variations, and is not intended for geometries substantially outside this range. This scope still exceeds that of conventional CXR segmentation methods, which are typically optimized for near-standard PA/AP images.

\endgroup

Additional limitations point to directions for future development. \textit{First}, the current projection pipeline relies on DRR based ray casting, which does not account for several physical effects present in real CXR, including scatter, beam hardening, detector blur, and antiscatter grid interactions. DRR rendering was chosen primarily for computational efficiency, enabling large scale data generation within practical time and hardware constraints. As simulation tools and computing resources advance, incorporating more physically realistic projection models, including Monte Carlo or hybrid approaches, may allow AnyCXR to more closely approximate clinical imaging conditions and further enhance generalization. \textit{Second}, the segmentation backbone within the CAR framework remains deliberately simple. We employed a UNet with residual blocks to isolate and evaluate the contributions of MSDR and CAR, yet even this lightweight model achieves strong cross view performance. Since the segmentation module is fully interchangeable, future work will examine newer architectures such as transformer based encoders and diffusion based segmentation models, which may improve boundary precision and robustness in anatomically complex regions. \textit{Third}, AnyCXR is trained entirely on synthetic data. While the model exhibits excellent zero shot generalization to clinical radiographs, synthetic only training does not expose the network to the full spectrum of acquisition artifacts, disease induced morphological changes, or scanner specific variations encountered in practice. From another perspective, AnyCXR can be interpreted as an anatomy focused foundation model for CXR, with strong structural recognition across diverse viewing geometries. A promising direction for future research involves using this model as an initialization for downstream tasks and exploring how limited real data can be used to adapt the model to specialized clinical applications.

%% file: conclusion.tex
\section{Conclusion}
\begingroup
We presented AnyCXR, a unified framework that enables comprehensive and generalizable anatomical segmentation for chest radiographs acquired at clinically relevant acquisition geometries. By integrating a large-scale Multi-stage Domain Randomization engine with the Conditional Joint Annotation Regularization framework, our method learns effectively from synthetic, partially labeled data and achieves strong zero-shot performance on real clinical images. The resulting anatomical maps consistently support downstream tasks, including cardiothoracic ratio estimation, spine curvature assessment, and disease classification, demonstrating the practical value of anatomically informed CXR analysis. Overall, AnyCXR offers a scalable foundation for future CXR-based AI systems, and continued advances in projection simulation and model architectures may further enhance its clinical applicability.
\endgroup

%% file: Appendix.tex
\appendix
\input{appendix_A}

\input{appendix_B}

\clearpage  

\onecolumn
\setlength{\floatsep}{6pt plus 2pt minus 2pt}
\setlength{\textfloatsep}{8pt plus 2pt minus 2pt}
\setlength{\intextsep}{6pt plus 2pt minus 2pt}
\captionsetup{skip=3pt}

\section{Detailed Distribution of Labels}
\label{app:C}
\begin{figure}[H]
    \centering
    \includegraphics[width=0.72\textwidth]{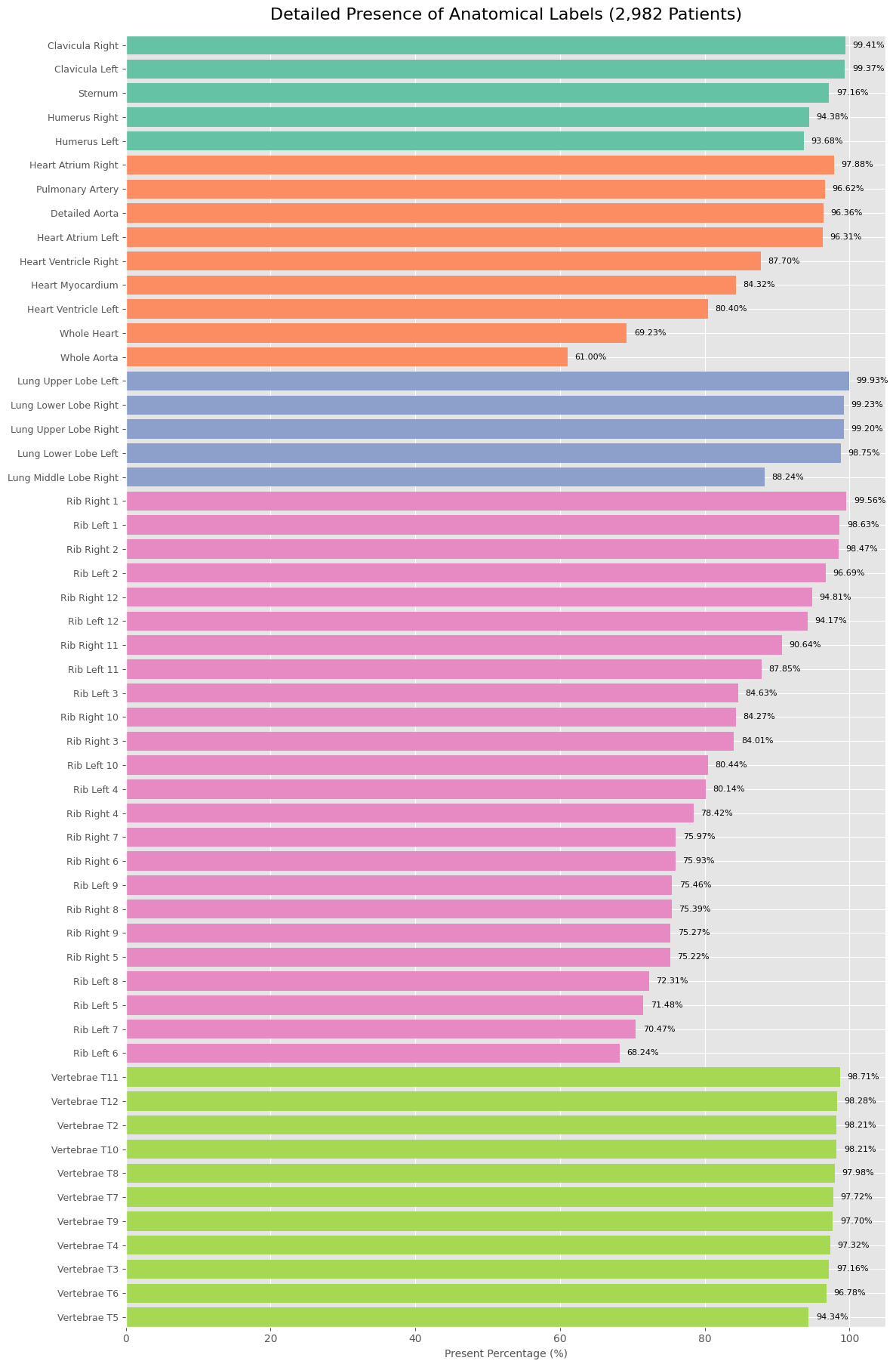}
    \captionsetup{justification=centering, singlelinecheck=false}
    \caption{Detailed distribution of each organ after filtering}
    \label{fig:dist2}
\end{figure}
\clearpage

\section{Segmentation Comparison of DRR Images in Oblique View} 
\label{app:D}
\begin{figure}[H]
    \centering
    \includegraphics[width=0.8\textwidth]{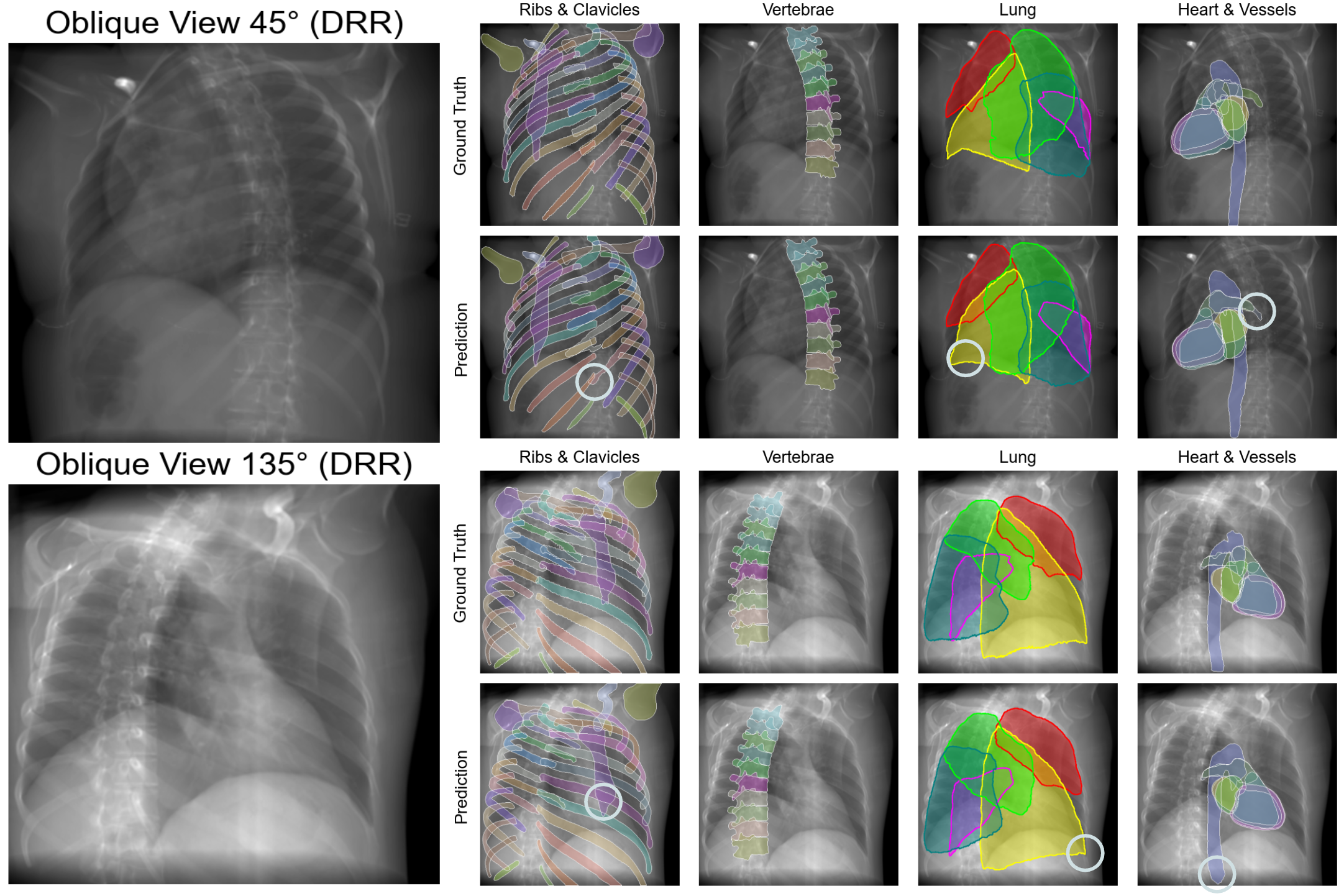}
    \captionsetup{justification=centering, singlelinecheck=false}
    \caption{Segmentation Comparison on Synthetic Oblique View DRRs}
    \label{fig:OBDRR}
\end{figure}

\section{Segmentation Performance of Oblique View} 
\label{app:E}
\input{seg_full_OB_TS}
\clearpage

\begingroup
\section{Segmentation Comparison Between FluoroSAM and AnyCXR}

\label{app:G}
\begin{figure}[!htbp]
    \centering
    \includegraphics[width=0.72\textwidth]{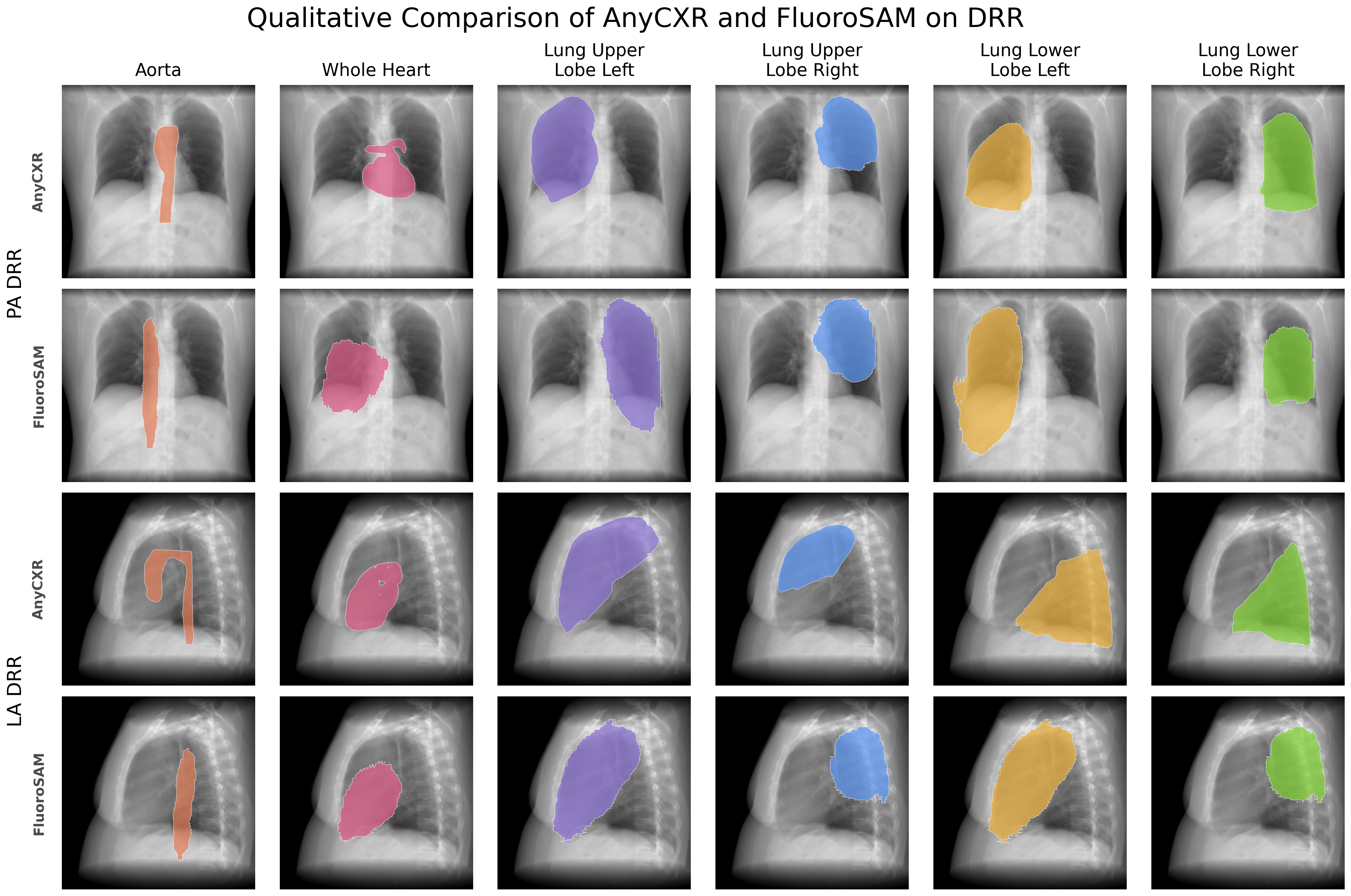}
    \captionsetup{justification=centering, singlelinecheck=false}
    \caption{Segmentation comparison of FluoroSAM and AnyCXR on synthetic DRR images in PA and LA views.}
    \label{fig:Fsam_Drr}
\end{figure}

\section{Segmentation Performance of FluoroSAM}
\label{app:H}
\input{FluoroSAM}

\clearpage

\section{Segmentation Comparison Between Rad-DINO and AnyCXR}
\label{app:I}
\begin{figure}[!htbp]
    \centering
    \includegraphics[width=0.72\textwidth]{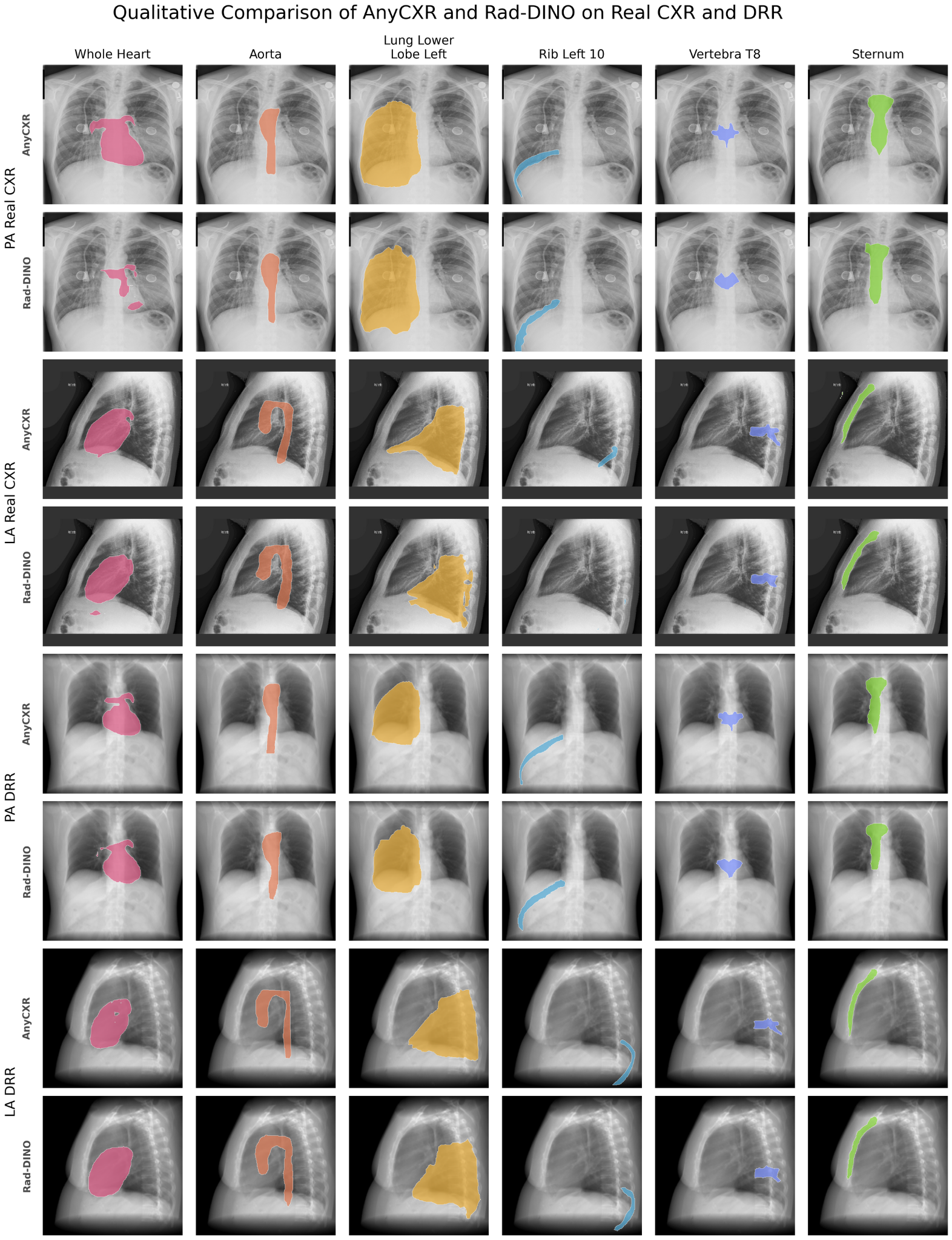}
    \captionsetup{justification=centering, singlelinecheck=false}
    \caption{
    Segmentation comparison of Rad-DINO and AnyCXR on real CXR and synthetic DRR images in PA and LA views.
    }
    \label{fig:Rad-DINO}
\end{figure}
\clearpage

\section {Segmentation Performance of Rad-DINO} 
\label{app:J}
\input{Rad-DINO}

\section{Statistical Significance of CAR Improvement}

\label{app:K}

Statistical analysis further supports the effectiveness of the proposed framework. On the CT-RATE DRR test set, all prespecified paired comparisons showed statistically significant improvements in Overall DSC (\(p < 0.001\)). Compared with the Plain baseline, MSDR Full improved DSC in both PA and LA views. MSDR Full also outperformed MSDR V1, confirming the added value of combining the augmentation components. Finally, adding CAR on top of MSDR Full produced a smaller but still significant gain in both views, with a larger effect observed in the LA view. Detailed baseline DSC, compared DSC, absolute gain, and 95\% confidence intervals are reported in Table~\ref{tab:paired_dsc_statistics}.
\input{stats}

\section {Ablation Study of HD for LA and PA Views} 
\label{app:F}
\input{ablation_study_HD}
\endgroup

\twocolumn

%% file: appendix_A.tex
\section{MSDR Implementation Details} 
\label{app:A}

In this section, we detail the mathematical formulations and hyperparameters used in the MSDR engine. The pipeline consists of 3D pre-projection augmentations followed by 2D post-projection detector simulations. We apply the following augmentations, each with uniform probability to individual masks.

\subsection*{\textbf{Notation and Mask Definitions}}
We define the volumetric segmentation masks used to isolate specific anatomical structures for targeted augmentation. Let $\Omega \subset \mathbb{R}^3$ denote the spatial domain of the CT volume. We define the set of all anatomical masks as $\mathcal{M}_{\text{ROI}} = M_s \cup M_b$, where:

\begin{itemize}
    \item \textbf{Soft Tissue ($M_s$):} The combined mask for all soft tissue structures.
    \item \textbf{Bone ($M_b$):} The global mask for all osseous structures.
\end{itemize}

It is explicitly noted that augmentations are restricted to voxels $x \in \mathcal{M}_{\text{ROI}}$. Voxels belonging to the background (air) or unmasked regions, $x \in \Omega \setminus \mathcal{M}_{\text{ROI}}$, remain invariant throughout the 3D augmentation pipeline.

Additionally, we define specific subsets of the bone mask:
\begin{itemize}
    \item \textbf{Vertebrae:} $\mathcal{V} = \{V_{T2}, V_{T3}, \dots, V_{T12}\}$.
    \item \textbf{Ribs:} $\mathcal{R}_L = \{R_{L1}, \dots, R_{L12}\}$ and $\mathcal{R}_R = \{R_{R1}, \dots, R_{R12}\}$.
\end{itemize}

\subsection*{\textbf{3D Attenuation Augmentations}}
Here, we detail the volumetric operations applied to modify the Hounsfield Units (HU) of the anatomical structures prior to the generation of the DRR.

\textbf{Bone-to-soft-tissue attenuation randomization}
This augmentation is applied with probability $p=0.4$. We apply anti-correlated scaling factors to $M_s$ and $M_b$ to disrupt consistent contrast relationships.
We define two range intervals: $S_{\text{high}} = [1.0, 1.7]$ and $S_{\text{low}} = [0.3, 1.0]$ and we randomly assign a tissue type $T_1 \in \{M_s, M_b\}$ to sample from $S_{\text{high}}$ and the other $T_2$ to sample from $S_{\text{low}}$. Let $s_1 \sim \mathcal{U}(S_{\text{high}})$ and $s_2 \sim \mathcal{U}(S_{\text{low}})$.

The voxel intensities $H(x)$ are updated as:
\begin{equation}
    H'(x) = 
    \begin{cases} 
    H(x) \cdot s_1 & \text{if } x \in T_1 \\
    H(x) \cdot s_2 & \text{if } x \in T_2
    \end{cases}
\end{equation}

\textbf{Inter-bone attenuation randomization}
We introduce two augmentation strategies to enhance bone density heterogeneity. The first decouples intensity correlations between adjacent bone structures, while the second simulates longitudinal attenuation gradients caused by varying tissue thickness and anatomical positioning.

\begin{itemize}

\item \textbf{Random Component Scaling:}
We apply HU value scaling for individual bone components, $\mathcal{K} = \mathcal{V} \cup \mathcal{R}_L \cup \mathcal{R}_R$ with probability $p=0.3$. For each distinct bone mask $M_k \in \mathcal{K}$, a unique scaling factor $s_k \sim \mathcal{U}(0.3, 1.7)$ is sampled. The intensity for voxels belonging to that specific bone is updated as:
\begin{equation}
    H'(x) = H(x) \cdot s_k, \quad \forall x \in M_k
\end{equation}

\item \textbf{Vertical Gradient Scale:}
We apply HU scaling along the vertical $Z$-axis (head-to-foot) to the spine ($\bigcup \mathcal{V}$), left ribs ($\bigcup \mathcal{R}_L$), and right ribs ($\bigcup \mathcal{R}_R$) independently with $p=0.4$. For each group, a normalized coordinate $\hat{z} \in [0, 1]$ spans the region from superior ($\hat{z}=0$) to inferior ($\hat{z}=1$). We sample $S_{\text{head}} \in [0.9, 1.5]$, $S_{\text{foot}} \in [0.6, 0.9]$, and $\alpha \in [0.5, 1.5]$ to define the scaling factor:
\begin{equation}
    S(\hat{z}) = S_{\text{foot}} + (1 - \hat{z})^\alpha \cdot (S_{\text{head}} - S_{\text{foot}})
\end{equation}
This factor is applied uniformly to all bone voxels within a given slice $z$.

\end{itemize}

\textbf{Intra-bone attenuation randomization}
We apply a morphology-based gradient to enhance the variation for the density transition from the cortical surface to the trabecular core with probability  $p=0.7$. We define depth via iterative erosion to handle irregular bone shapes. 

Let $\delta(x) \in [0, 1]$ denote the normalized morphological depth of voxel $x$, where $\delta=0$ corresponds to the outer surface and $\delta=1$ to the deepest core layer found by recursive binary erosion. We sample surface and core scaling factors $S_{\text{surf}} \sim \mathcal{U}(0.9, 1.5)$ and $S_{\text{core}} \sim \mathcal{U}(0.6, 1.1)$. The intensity update is defined as:
\begin{equation}
    H'(x) = H(x) \cdot \left[ S_{\text{surf}} + \delta(x)^\alpha \cdot (S_{\text{core}} - S_{\text{surf}}) \right]
\end{equation}
where $\alpha \in [0.5, 1.5]$ controls the gradient curvature.

\textbf{Soft-tissue attenuation randomization}
We apply one of two mutually exclusive augmentations to $M_s$.

\begin{itemize}
\item  \textbf{Standard Scaling ($p=0.6$):} 
We apply a uniform scalar $s \sim \mathcal{U}(0.3, 1.7)$ to each soft tissue voxel in $M_s$: $H'(x) = H(x) \cdot s$.

\item  \textbf{Polarity Inversion ($p=0.3$):} 
To simulate extreme contrast reversal, we map soft tissue values to a positive range. We sample a drift $\delta \sim \mathcal{U}(-0.1, 0.1)$ and apply the transformation:
\begin{equation}
    H'(x) = -(1.5 + \delta) \cdot H(x)
\end{equation}
This is applied to voxels within the valid soft tissue range (e.g., $HU \in [-900, -2]$).

\end{itemize}

\subsection*{\textbf{Geometric and Noise Augmentations}}
This section outlines the stochastic perturbations applied to the projection geometry parameters, as well as the injection of noise and synthetic foreign objects into the simulation environment.

\textbf{Noise and artifacts}

\begin{itemize}
    \item \textbf{Gaussian Noise ($p=0.3$):} Voxel-wise additive noise $N \sim \mathcal{N}(0, \sigma^2)$ is added to the volume, where $\sigma$ is uniformly sampled per volume from $[10, 50]$ HU.
    \item \textbf{External Objects ($p=0.2$):} We inject $N \in \{1, 2, 3\}$ synthetic objects (cylinders or ellipsoids). Dimensions are sampled uniformly (Radius $\sim \mathcal{U}(5, 20)$ mm, Length $\sim \mathcal{U}(10, 60)$ mm) and objects are assigned $HU \in (1800, 2000)$ to simulate dense foreign bodies.
\end{itemize}

\textbf{Projection geometry randomization}

\begin{itemize}
    \item \textbf{Source-to-Detector distance (SDD):} Applied with $p=0.4$. The SDD is perturbed by a factor $s_{sod} \sim \mathcal{U}(0.9, 1.1)$.
    \item \textbf{Object-to-Detector Distance (ODD):} Applied with $p=0.4$. A vertical offset $\Delta h \sim \mathcal{U}(-30, 30)$ mm is added to the ODD.
\end{itemize}

\subsection*{\textbf{2D Post-Projection Augmentations}}
Finally, we describe the image-level transformations applied to the projected 2D radiographs to simulate varying detector responses and exposure conditions.

\textbf{Exposure and Detector–response Normalization}
With probability $p=0.7$, we apply a piecewise-linear tone mapping defined by a randomized control point $(x'_c, y'_c)$. We sample the knot coordinates as $x'_c \sim \mathcal{U}[-0.2, 0.4]$ and $y'_c \sim \mathcal{U}(-0.2, 0.4)$. The final pixel intensity is obtained by linear interpolation between the anchors $(0,0)$, $(x'_c, y'_c)$, and $(1,1)$.

\textbf{Polarity perturbation}
With probability $p=0.3$, we invert the global intensity of the 2D image $I$:
\begin{equation}
    I_{\text{new}} = (I_{\min} + I_{\max}) - I + \epsilon
\end{equation}
where $\epsilon$ is a small stability term.

%% file: appendix_B.tex
\section{Model Implementation Details}
\label{app:B}
The segmentation backbone is constructed as a U-Net with a ResNet-50 encoder. We initialize the network using a standard ImageNet-pretrained ResNet-50 and adapt the first convolutional block ($\texttt{conv1}$) to handle the single-channel CXR inputs. The encoder stages expose the feature maps after $\texttt{conv1}+\texttt{bn1}+\texttt{relu}+\texttt{maxpool}$ and $\texttt{layer1}$ through $\texttt{layer4}$, yielding feature strides of $1/4$, $1/8$, $1/16$, and $1/32$ of the input resolution. Input CXR images are normalized to the range $[0, 1]$ prior to processing. The decoder mirrors these scales, using transposed convolutions and skip connections at each upsampling stage to reconstruct the final multi-class probability maps.

The CAR framework operates on the full multi-class probability maps without downsampling. The CAR encoder $G_{\theta_1}$ is conditioned by using direct channel-wise concatenation of the input image $\vect{X}$ and the prediction $\hat{\vect Y}$. The encoder consists of two convolutional blocks: a $3\times3$ convolution mapping from $C_{\text{in}}^{\text{CAR}}$ channels (where $C_{\text{in}}^{\text{CAR}}$ is $1$ plus the number of output classes) to $C_{\text{lat}}/4$ channels, followed by batch normalization, ReLU, and a single $4\times4$ max-pooling layer (resulting in overall $4\times$ spatial downsampling), and a second $3\times3$ convolution to $C_{\text{lat}}$ channels with batch normalization and ReLU. The CAR decoder $G_{\theta_2}$ is a lightweight symmetric stack comprising two $4\times4$ transposed convolutions (stride 2, padding 1) that upsample from $C_{\text{lat}}$ to $C_{\text{lat}}/2$ and then to $C_{\text{lat}}/4$, each followed by batch normalization and ReLU, and a final $1\times1$ convolution to the output channel dimension with sigmoid activation.

%% file: seg_full_OB_TS.tex
\begingroup
\begin{table*}[htbp]

\centering
\scriptsize
\renewcommand{\arraystretch}{1.15}
\setlength{\tabcolsep}{5pt}
\caption{Dice and Hausdorff Distance (mm) by Anatomical Structure for the Oblique View}
\label{tab:full_seg_oblique_TS}

\makebox[\linewidth][c]{%
\begin{tabular}{lcccc | lcccc}
\toprule
\textbf{Structure}
& \textbf{45$^\circ$ DSC$\uparrow$}
& \textbf{45$^\circ$ HD$\downarrow$}
& \textbf{135$^\circ$ DSC$\uparrow$}
& \textbf{135$^\circ$ HD$\downarrow$}
& \textbf{Structure}
& \textbf{45$^\circ$ DSC$\uparrow$}
& \textbf{45$^\circ$ HD$\downarrow$}
& \textbf{135$^\circ$ DSC$\uparrow$}
& \textbf{135$^\circ$ HD$\downarrow$} \\
\midrule
Vertebra T2 & 0.913 & 4.4 & 0.909 & 7.9 & Whole Heart & 0.905 & 20.7 & 0.921 & 21.2 \\
Vertebra T3 & 0.916 & 4.6 & 0.926 & 5.6 & Heart Atrium Left & 0.838 & 10.6 & 0.890 & 8.5 \\
Vertebra T4 & 0.936 & 4.1 & 0.919 & 5.6 & Heart Atrium Right & 0.825 & 14.4 & 0.894 & 10.1 \\
Vertebra T5 & 0.942 & 4.0 & 0.921 & 5.4 & Heart Myocardium & 0.940 & 6.3 & 0.931 & 8.7 \\
Vertebra T6 & 0.929 & 7.0 & 0.932 & 5.7 & Heart Ventricle Left & 0.926 & 7.7 & 0.921 & 7.7 \\
Vertebra T7 & 0.933 & 2.5 & 0.940 & 4.4 & Heart Ventricle Right & 0.869 & 15.3 & 0.917 & 9.5 \\
Vertebra T8 & 0.932 & 3.3 & 0.939 & 3.7 & Pulmonary Artery & 0.866 & 14.2 & 0.890 & 11.3 \\
Vertebra T9 & 0.912 & 8.1 & 0.944 & 5.0 & Aorta & 0.898 & 9.9 & 0.885 & 19.7 \\
Vertebra T10 & 0.923 & 6.3 & 0.939 & 5.6 & Lung & 0.925 & 17.5 & 0.935 & 19.5 \\
Vertebra T11 & 0.938 & 3.8 & 0.935 & 5.1 & Lung Upper Lobe Left & 0.928 & 20.1 & 0.932 & 18.5 \\
Vertebra T12 & 0.932 & 5.6 & 0.938 & 5.5 & Lung Lower Lobe Left & 0.924 & 19.6 & 0.944 & 15.5 \\
Clavicle Left & 0.923 & 10.1 & 0.928 & 6.2 & Lung Upper Lobe Right & 0.947 & 13.2 & 0.939 & 16.7 \\
Clavicle Right & 0.929 & 2.9 & 0.938 & 3.9 & Lung Middle Lobe Right & 0.888 & 19.5 & 0.897 & 17.8 \\
Humerus Left & 0.961 & 4.0 & 0.875 & 6.1 & Lung Lower Lobe Right & 0.925 & 18.6 & 0.928 & 23.4 \\
Humerus Right & 0.963 & 2.6 & 0.959 & 5.1 & Sternum & 0.905 & 16.0 & 0.917 & 11.4 \\
Rib Left 1 & 0.918 & 4.5 & 0.899 & 8.9 & Rib Right 1 & 0.910 & 3.9 & 0.906 & 5.3 \\
Rib Left 2 & 0.900 & 5.3 & 0.871 & 11.3 & Rib Right 2 & 0.880 & 6.1 & 0.902 & 8.4 \\
Rib Left 3 & 0.877 & 8.9 & 0.838 & 24.4 & Rib Right 3 & 0.919 & 4.4 & 0.911 & 5.4 \\
Rib Left 4 & 0.907 & 7.0 & 0.838 & 19.8 & Rib Right 4 & 0.911 & 7.1 & 0.920 & 5.3 \\
Rib Left 5 & 0.886 & 14.3 & 0.865 & 17.5 & Rib Right 5 & 0.910 & 13.7 & 0.926 & 6.4 \\
Rib Left 6 & 0.847 & 19.0 & 0.879 & 13.4 & Rib Right 6 & 0.882 & 22.5 & 0.932 & 5.0 \\
Rib Left 7 & 0.900 & 13.7 & 0.903 & 10.3 & Rib Right 7 & 0.913 & 12.9 & 0.934 & 5.7 \\
Rib Left 8 & 0.923 & 13.7 & 0.885 & 13.3 & Rib Right 8 & 0.916 & 9.1 & 0.914 & 8.8 \\
Rib Left 9 & 0.933 & 7.3 & 0.889 & 14.0 & Rib Right 9 & 0.918 & 4.8 & 0.909 & 11.4 \\
Rib Left 10 & 0.821 & 21.7 & 0.918 & 9.9 & Rib Right 10 & 0.893 & 19.8 & 0.923 & 12.6 \\
Rib Left 11 & 0.768 & 22.5 & 0.932 & 4.4 & Rib Right 11 & 0.835 & 23.5 & 0.922 & 5.1 \\
Rib Left 12 & 0.835 & 13.7 & 0.912 & 4.7 & Rib Right 12 & 0.911 & 2.6 & 0.887 & 5.0 \\
\bottomrule
\end{tabular}
}% end makebox

\end{table*}
\endgroup

%% file: FluoroSAM.tex
\begingroup
\begin{table*}[htb]

\centering
\scriptsize
\renewcommand{\arraystretch}{1.20}
\setlength{\tabcolsep}{6pt}
\caption{FluoroSAM segmentation statistics by organ group on synthetic DRR (PA and LA views)}
\label{tab:fsam_drr}

\makebox[\linewidth][c]{%
\begin{tabular}{lcccc | lcccc}
\toprule
\textbf{Group: PA} 
& \textbf{DSC$\uparrow$}
& \textbf{Std DSC}
& \textbf{HD$\downarrow$}
& \textbf{Std HD}
& \textbf{Group: LA} 
& \textbf{DSC$\uparrow$}
& \textbf{Std DSC}
& \textbf{HD$\downarrow$}
& \textbf{Std HD} \\
\midrule

Heart \& Vessels
& 0.132 & 0.177 & 98.5 & 47.3
& Heart \& Vessels
& 0.226 & 0.262 & 95.5 & 55.9 \\

Lung Group
& 0.600 & 0.304 & 73.6 & 54.5
& Lung Group
& 0.644 & 0.230 & 75.6 & 34.8 \\

Rib Group
& 0.124 & 0.156 & 144.9 & 98.5
& Rib Group
& 0.105 & 0.148 & 109.0 & 68.6 \\

Vertebrae Group
& 0.421 & 0.283 & 44.8 & 51.1
& Vertebrae Group
& 0.297 & 0.273 & 56.8 & 55.8 \\

Other Bone Group
& 0.376 & 0.253 & 85.5 & 91.5
& Other Bone Group
& 0.394 & 0.191 & 63.1 & 37.0 \\

\bottomrule
\end{tabular}
}% end makebox

\end{table*}

\begin{table*}[htb]

\centering
\scriptsize
\renewcommand{\arraystretch}{1.20}
\setlength{\tabcolsep}{6pt}
\caption{FluoroSAM segmentation statistics by organ group on expert-refined real CXR (PA and LA views)}
\label{tab:fsam_real}

\makebox[\linewidth][c]{%
\begin{tabular}{lcccc | lcccc}
\toprule
\textbf{Group: PA} 
& \textbf{DSC$\uparrow$}
& \textbf{Std DSC}
& \textbf{HD$\downarrow$}
& \textbf{Std HD}
& \textbf{Group: LA} 
& \textbf{DSC$\uparrow$}
& \textbf{Std DSC}
& \textbf{HD$\downarrow$}
& \textbf{Std HD} \\
\midrule

Heart \& Vessels
& 0.150 & 0.168 & 148.8 & 79.2
& Heart \& Vessels
& 0.122 & 0.160 & 340.7 & 305.4 \\

Lung Group
& 0.209 & 0.208 & 227.7 & 111.7
& Lung Group
& 0.252 & 0.249 & 414.5 & 369.6 \\

Rib Group
& 0.010 & 0.046 & 313.7 & 153.4
& Rib Group
& 0.024 & 0.081 & 606.6 & 546.4 \\

Vertebrae Group
& 0.031 & 0.112 & 180.6 & 101.6
& Vertebrae Group
& 0.000 & 0.000 & 672.0 & 536.4 \\

Other Bone Group
& 0.045 & 0.086 & 355.6 & 196.4
& Other Bone Group
& 0.000 & 0.000 & 1016.0 & 777.9 \\

\bottomrule
\end{tabular}
}% end makebox

\end{table*}
\endgroup

%% file: Rad-DINO.tex
\begingroup
\begin{table*}[!htbp]

\centering
\scriptsize
\renewcommand{\arraystretch}{1.20}
\setlength{\tabcolsep}{6pt}
\caption{Rad-DINO segmentation statistics by organ group on synthetic DRR (PA and LA views)}
\label{tab:raddino_drr_car_on}

\makebox[\linewidth][c]{%
\begin{tabular}{lcccc | lcccc}
\toprule
\textbf{Group: PA}
& \textbf{DSC$\uparrow$}
& \textbf{Std DSC}
& \textbf{HD$\downarrow$}
& \textbf{Std HD}
& \textbf{Group: LA}
& \textbf{DSC$\uparrow$}
& \textbf{Std DSC}
& \textbf{HD$\downarrow$}
& \textbf{Std HD} \\
\midrule

Heart \& Vessels
& 0.892
& 0.143
& 4.6
& 4.4
& Heart \& Vessels
& 0.885
& 0.164
& 5.8
& 8.0 \\

Lung Group
& 0.934
& 0.073
& 8.3
& 7.7
& Lung Group
& 0.926
& 0.033
& 11.2
& 6.6 \\

Rib Group
& 0.747
& 0.233
& 7.3
& 12.3
& Rib Group
& 0.623
& 0.268
& 12.5
& 18.7 \\

Vertebrae Group
& 0.874
& 0.076
& 4.0
& 2.0
& Vertebrae Group
& 0.865
& 0.046
& 3.8
& 1.1 \\

Other Bone Group
& 0.869
& 0.133
& 5.7
& 9.5
& Other Bone Group
& 0.765
& 0.278
& 7.2
& 6.1 \\

\bottomrule
\end{tabular}
}%
\end{table*}

\begin{table*}[htbp!]

\centering
\scriptsize
\renewcommand{\arraystretch}{1.20}
\setlength{\tabcolsep}{6pt}
\caption{Rad-DINO segmentation statistics by organ group on expert-refined real CXR (PA and LA views)}
\label{tab:raddino_real}

\makebox[\linewidth][c]{%
\begin{tabular}{lcccc | lcccc}
\toprule
\textbf{Group: PA}
& \textbf{DSC$\uparrow$}
& \textbf{Std DSC}
& \textbf{HD$\downarrow$}
& \textbf{Std HD}
& \textbf{Group: LA}
& \textbf{DSC$\uparrow$}
& \textbf{Std DSC}
& \textbf{HD$\downarrow$}
& \textbf{Std HD} \\
\midrule

Heart \& Vessels
& 0.877
& 0.205
& 22.4
& 6.4
& Heart \& Vessels
& 0.872
& 0.250
& 30.4
& 11.3 \\

Lung Group
& 0.914
& 0.100
& 40.1
& 11.6
& Lung Group
& 0.898
& 0.043
& 57.0
& 10.0 \\

Rib Group
& 0.726
& 0.351
& 37.9
& 17.2
& Rib Group
& 0.617
& 0.406
& 61.3
& 26.8 \\

Vertebrae Group
& 0.866
& 0.111
& 19.0
& 2.8
& Vertebrae Group
& 0.855
& 0.071
& 19.7
& 1.7 \\

Other Bone Group
& 0.852
& 0.209
& 28.1
& 14.3
& Other Bone Group
& 0.756
& 0.419
& 37.6
& 9.2 \\

\bottomrule
\end{tabular}
}% end makebox

\end{table*}
\endgroup

%% file: stats.tex
\begingroup
\begin{table*}[htbp!]

\centering
\scriptsize
\renewcommand{\arraystretch}{1.20}
\setlength{\tabcolsep}{6pt}
\caption{Paired statistical comparison of Overall DSC improvements on CT-RATE DRR test set}
\label{tab:paired_dsc_statistics}

\makebox[\linewidth][c]{%
\begin{minipage}{\linewidth}
\centering
\begin{tabular}{llccccc}
\toprule
\textbf{Comparison}
& \textbf{View}
& \textbf{Baseline DSC}
& \textbf{Compared DSC}
& \textbf{Gain}
& \textbf{95\% CI Lower}
& \textbf{95\% CI Upper} \\
\midrule

Plain vs. MSDR Full
& PA
& 90.86
& 94.13
& +3.27
& 2.84
& 3.69 \\

Plain vs. MSDR Full
& LA
& 83.39
& 89.90
& +6.51
& 5.09
& 7.93 \\

MSDR V1 vs. MSDR Full
& PA
& 93.75
& 94.13
& +0.38
& 0.27
& 0.49 \\

MSDR V1 vs. MSDR Full
& LA
& 86.94
& 89.90
& +2.96
& 1.71
& 4.21 \\

MSDR Full vs. Full + CAR
& PA
& 94.13
& 94.56
& +0.43
& 0.37
& 0.49 \\

MSDR Full vs. Full + CAR
& LA
& 89.90
& 90.71
& +0.81
& 0.44
& 1.18 \\

\bottomrule
\end{tabular}

\vspace{3pt}
{\scriptsize \textit{Notes.} Confidence intervals are 95\% paired \(t\)-intervals over structure-level DSC differences; all paired Wilcoxon signed-rank tests were significant at \(p < 0.001\).}
\end{minipage}
}%

\end{table*}
\endgroup

%% file: ablation_study_HD.tex
\begingroup

\centering
\footnotesize
\captionof{table}{Ablation study of HD for major views and organ sub-groups. Lower values indicate better performance. Within each column, the best-performing method is highlighted in bold.}
\label{tab:ablation_HD_relayout}
\makebox[\linewidth][c]{%
\begin{minipage}{\linewidth}
\centering
\setlength{\tabcolsep}{4pt}
\renewcommand{\arraystretch}{1.15}

\begin{tabular}{llcccccc}
\toprule
\textbf{View} & \textbf{Method} & \textbf{Overall$\downarrow$} & \textbf{Heart \& Vessels$\downarrow$} & \textbf{Lung$\downarrow$} & \textbf{Rib$\downarrow$} & \textbf{Vertebrae$\downarrow$} & \textbf{Other Bone$\downarrow$} \\
\midrule
PA & Plain
& $11.04 \pm 10.75$
& $11.12 \pm 9.40$
& $17.90 \pm 6.98$
& $10.35 \pm 12.59$
& $7.77 \pm 7.99$
& $8.05 \pm 14.23$ \\
PA & MSDR V1
& $\mathbf{6.92 \pm 5.60}^{\dagger}$
& $8.34 \pm 4.03^{\circ}$
& $15.35 \pm 6.65^{*}$
& $5.96 \pm 7.40^{\dagger}$
& $4.81 \pm 2.73^{\ddagger}$
& $5.08 \pm 5.10$ \\
PA & MSDR V2
& $6.94 \pm 7.83^{\dagger}$
& $\mathbf{7.27 \pm 5.12}^{\circ}$
& $\mathbf{12.53 \pm 8.41}^{*}$
& $7.18 \pm 11.22^{\dagger}$
& $4.79 \pm 3.88^{\ddagger}$
& $4.33 \pm 4.54^{*}$ \\
PA & MSDR Full
& $7.08 \pm 5.69^{\dagger}$
& $7.39 \pm 3.94^{\circ}$
& $13.54 \pm 7.21^{*}$
& $\mathbf{5.75 \pm 7.83}^{\dagger}$
& $\mathbf{4.48 \pm 3.35}^{\ddagger}$
& $\mathbf{4.25 \pm 2.29}^{*}$ \\
\midrule
LA & Plain
& $16.46 \pm 12.95$
& $10.62 \pm 8.25$
& $22.81 \pm 9.94$
& $25.38 \pm 19.10$
& $8.40 \pm 4.66$
& $15.11 \pm 13.13$ \\
LA & MSDR V1
& $14.33 \pm 8.73^{\dagger}$
& $\mathbf{9.48 \pm 5.72}^{*}$
& $20.63 \pm 9.13$
& $22.21 \pm 16.46^{\circ}$
& $5.89 \pm 2.23^{\ddagger}$
& $13.44 \pm 10.14$ \\
LA & MSDR V2
& $14.25 \pm 12.19^{\dagger}$
& $13.01 \pm 13.20$
& $17.95 \pm 7.92$
& $18.29 \pm 17.01^{\dagger}$
& $6.53 \pm 3.77^{\circ}$
& $10.42 \pm 10.03$ \\
LA & MSDR Full
& $\mathbf{10.94 \pm 8.40}^{\dagger}$
& $10.33 \pm 7.95$
& $\mathbf{17.42 \pm 8.91}$
& $\mathbf{13.96 \pm 12.11}^{\dagger}$
& $\mathbf{5.05 \pm 1.85}^{\ddagger}$
& $\mathbf{7.95 \pm 5.20}$ \\
\bottomrule
\end{tabular}%

\vspace{0.4em}

{\footnotesize
\textit{Notes.} 
Statistical significance was assessed against Plain using a two-sided paired Wilcoxon signed-rank test within the same row: $^{*}p < 0.05$, $^{\circ}p < 0.01$, $^{\ddagger}p < 0.001$, $^{\dagger}p < 0.0001$.
}
\end{minipage}
}%
\endgroup